%% file: main.tex
\documentclass[runningheads]{llncs}
\usepackage{multirow}
\usepackage[T1]{fontenc}
\usepackage{float} 
\usepackage{subcaption}
\input{mathcommands.tex}

\input{cmds}
\usepackage{booktabs}
\usepackage{graphicx}
\usepackage{svg}
\usepackage{xargs}
\usepackage{mathtools}

\begin{document}
\title{Which Direction to Choose? An Analysis on the Representation Power of Self-Supervised ViTs in Downstream Tasks}
\author{Yannis Kaltampanidis\inst{1}\orcidID{0009-0008-9624-7783} \and \\
Alexandros Doumanoglou\inst{1}\orcidID{0000-0002-4337-1720} \and \\
Dimitrios Zarpalas \inst{1}\orcidID{0000-0002-9649-9306}}
\authorrunning{Y. Kaltampanidis et al.}
\titlerunning{Which Direction to Choose ?}
\institute{Information Technologies Institute, Centre for Research and Technology Hellas, 1st Km Charilaou - Thermi Road, Thessaloniki, Greece \\ 
\email{\{ykalt, aldoum, zarpalas\}@iti.gr}\\
\url{https://www.iti.gr}}
\maketitle              %
\begin{abstract}
Self-Supervised Learning (SSL) for Vision Transformers (ViTs) has recently demonstrated considerable potential as a pre-training strategy for a variety of computer vision tasks, including image classification and segmentation, both in standard and few-shot downstream contexts. Two pre-training objectives dominate the landscape of SSL techniques: Contrastive Learning and Masked Image Modeling. Features (or tokens) extracted from the final transformer attention block --specifically, the keys, queries, and values-- as well as features obtained after the final block's feed-forward layer, have become a common foundation for addressing downstream tasks. However, in many existing approaches, these pre-trained ViT features are further processed through additional transformation layers, often involving lightweight heads or combined with distillation, to achieve superior task performance. Although such methods can improve task outcomes, to the best of our knowledge, a comprehensive analysis of the intrinsic representation capabilities of unaltered ViT features has yet to be conducted. This study aims to bridge this gap by systematically evaluating the use of these unmodified features across image classification and segmentation tasks, in both standard and few-shot contexts. The classification and segmentation rules that we use are either hyperplane based (as in logistic regression) or cosine-similarity based, both of which rely on the presence of interpretable directions in the ViT's latent space. Based on the previous rules and without the use of additional feature transformations, we conduct an analysis across token types, tasks, and pre-trained ViT models. This study provides insights into the optimal choice for token type and decision rule based on the task, context, and the pre-training objective, while reporting detailed findings on two widely-used datasets.

\keywords{ViT  \and SSL \and DiNO \and MAE \and Directions \and Hyperplane \and Cosine Similarity}
\end{abstract}
\section{Introduction}

\input{sections/introduction}
\input{sections/related_work}

\input{sections/approach}

\input{sections/results}

\input{sections/conclusion}

\bibliography{refs}
\bibliographystyle{splncs04}
\end{document}

%% file: mathcommands.tex
\usepackage{amsmath,amsfonts,bm}

\def\eqref#1{equation~\ref{#1}}

\def\1{\bm{1}}

\def\vd{{\bm{d}}}

\def\vk{{\bm{k}}}

\def\vq{{\bm{q}}}

\def\vv{{\bm{v}}}
\def\vw{{\bm{w}}}
\def\vx{{\bm{x}}}

\def\vz{{\bm{z}}}

\DeclareMathAlphabet{\mathsfit}{\encodingdefault}{\sfdefault}{m}{sl}
\SetMathAlphabet{\mathsfit}{bold}{\encodingdefault}{\sfdefault}{bx}{n}

%% file: cmds.tex
\def\k{\vk}
\def\q{\vq}
\def\v{\vv}
\def\z{\vz}
\def\d{\vd}
\def\x#1{\bm{x}_{#1}}
\def\w{\vw}
\def\valpha{{\bm{\alpha}}}

%% file: sections/introduction.tex
Vision transformers \cite{vit,deit,vit-survey1,vit-survey2,vit-survey3}, have shown exceptional performance in addressing complex computer vision and multi-modal tasks \cite{pali,beit3,Radford2021LearningTV,one_peace}. However, their effectiveness is highly dependent on the size of the training dataset, requiring an extensive amount of data to generalize effectively and avoid overfitting. Training these models from scratch is resource-intensive, both in terms of computational power and processing time. Given that related tasks, such as classification and segmentation, often share foundational knowledge, training separate models for each task from scratch is inefficient. Therefore, it has been proposed to train a large model once, using substantial data and resources to capture general knowledge, and then specialize or distill this model for specific downstream tasks by leveraging the knowledge acquired during the initial training phase.

Self-supervision, based on Masked Image Modeling (MIM) \cite{mae,beit,simclr} or Contrastive Learning (CL) \cite{dino,moco,simclr}, has been proposed as a way for ViTs to capture this general knowledge from large datasets without the need for explicit labels. However, to achieve top performance, in most approaches \cite{tokencut,cutler,stego,enchancer} the pre-trained ViT features undergo further transformations before the final prediction, in order to align the feature representations with the solution of the downstream task. Moreover, different methods utilize various feature types --such as query-key-value pairs from the last attention block, or the output tokens of the final feed forward layer-- and employ diverse decision rules, being either hyperplane-based or direction similarity-based. Even though these approaches have demonstrated their effectiveness in solving downstream tasks, yet to our knowledge, a comprehensive evaluation of the intrinsic representation capabilities of unaltered self-supervised ViT features is missing from the literature.

In this work, we present a comprehensive analysis of the representational power of unaltered features from two self-supervised ViTs, pre-trained on a large dataset \cite{imagenet} using the previously mentioned self-supervision objectives \cite{mae,dino}. To the best of our knowledge, this is the first study to examine all of the following aspects simultaneously: a) two ViTs pre-trained with different \textbf{self-supervised objectives} b) the five possible \textbf{token types} from the last transformer layer --keys, queries, values, and features before and after the final feed-forward block-- c) two downstream \textbf{tasks}: image classification and segmentation, across both standard and 1-way-k-shot \textbf{contexts} and d) two commonly used prediction methods (or, as otherwise mentioned, \textbf{decision rules}), based on either hyperplane separation (linear probing) or cosine similarity. 
\par 
We find that the hyperplane decision rule is more effective in semantic separability across most experiments, indicating that the cosine similarity between the tokens of these pretrained models is a suboptimal semantic proximity metric. Furthermore, our experiments indicate that the optimal token type depends heavily on the pre-training objective, task, context and decision rule --with some previously overlooked tokens proving to be the most effective. Beyond practical guidelines, our work challenges existing intuitions about ViT token interpretations and underscores the need for a deeper understanding of the role of each computational block within ViT layers.

%% file: sections/related_work.tex
\section{Related Work}

\textbf{Self-Supervised Pre-Training} Self-supervised pre-training \cite{self_supervised} stands out as the leading method towards developing vision, vision-language, and various multi-modal foundation models \cite{foundation_models,one_peace,coca}. The core strategies in this field involve CL, MIM, or an integration of both. On the one hand, CL methods \cite{simclr,dense_clr,moco_1,moco} utilize image augmentation techniques to generate views with similar or dissimilar semantic content, which, in turn, are considered for feature alignment. On the other hand, representation learning in MIM methods \cite{beit,beit2,beit3} is driven by masking patches and then reconstructing pixels or features. 

Within ViTs, MIM approaches, largely represented by Masked Autoencoders (MAEs) \cite{mae,simim}, typically \textbf{require} supervised \textbf{fine-tuning} to achieve competitive performance on downstream tasks \cite{mae,beit,simim,Zhang2023MultimodalFA,mae-medical,Nguyen2024ExploringSV}. These models tend to exhibit narrow self-attention receptive fields \cite{Yue2023UnderstandingMA} and capture texture-based features, making them best suited for dense prediction tasks such as object detection \cite{Park2023WhatDS}. They also tend to exhibit great scaling with an increasing number of parameters which can be attributed to the high attention-map variance between transformer heads, meaning that a larger portion of the network can being utilized during fine-tuning \cite{Park2023WhatDS}.

ViTs trained with a CL framework, such as DiNO \cite{dino}, generate semantic-level feature representations \cite{Amir2021DeepVF}, allowing them to serve as universal feature extractors \textbf{without further fine-tuning} \cite{Vanyan2023AnalyzingLR}. Similar to other contrastive learning methods, the self-attention maps of a ViT pre-trained with a DiNO objective, have a broad receptive field, effectively capturing global patterns, but CL also faces the challenge of \textit{collapse into homogeneity} \cite{Park2023WhatDS}, leading to similar self-attention maps for all heads. This limitation has motivated the development of hybrid SSL techniques that combine MIM and CL learning objectives to address their respective limitations \cite{dinov2,Park2023WhatDS,Mishra2022ASE,mae-probe}.

\noindent \textbf{Transfer-Learning Self-Supervised ViTs on Downstream Tasks} In dense prediction tasks, the \textit{patch} tokens of the final encoder layer are commonly used as regional embeddings \cite{one_peace,Yang2024DepthAU,mae}, while the corresponding \textit{class} token ({[CLS]}) remains the standard representation for image classification \cite{vit,dino,Zhang2023MultimodalFA}. The ability of DiNO to induce discriminative saliency maps in the self-attention mechanism of ViTs \cite{dino} has inspired the extraction of features directly from the self-attention blocks. Beyond the vanilla approach that uses the class token for image classification tasks, various techniques have been explored that leverage the \textit{key} tokens in the self-attention block of a frozen DiNO backbone (a ViT pre-trained with the DiNO objective), to tackle unsupervised segmentation and localization tasks \cite{Lost,tokencut,cutler}, often employing a cosine similarity-based signal. Alternative methods that utilize a similar backbone seek to distill its knowledge in both standard \cite{stego} and few-shot \cite{enchancer} contexts through lightweight heads, using the backbone as a means to detect semantic similarities within the data.

Unlike CL which is able to build strong frozen backbones, MIM pre-training is best capitalized with task-specific fine-tuning. In \cite{Zhang2023MultimodalFA}, a MAE is pre-trained on a face dataset and subsequently fine-tuned on a dataset with facial expressions for facial affective analysis. In the medical domain, where annotated data are more scarce, self-pre-training \cite{mae-medical} has been proposed as the paradigm of pre-training a MAE directly on the data of the downstream task. Subsequently, the learned encoder can be combined with a trainable linear head or a convolutional decoder to demonstrate superior performance compared to supervised baselines or baselines pre-trained on out-of-domain data. Beyond masking pixels, the MIM objective can be utilized to train lightweight student models that learn to reconstruct masked features from a larger state-of-the-art teacher, providing efficient solutions to solve the downstream segmentation task \cite{efficientsam}.

\noindent \textbf{Relation to the Present Work} In contrast to other studies that shed light on self-supervised ViTs from varying perspectives \cite{moco,Park2023WhatDS,Yue2023UnderstandingMA,vit-see,vit-intriguing}, our research adopts a latent space probing approach, regularly explored in mechanistic interpretability \cite{Zhou2018InterpretableBD,doumanoglou2023unsupervised,fel2023craft,cunningham2023sparse,rao2024discover}. 
To our knowledge, this study is the first to rigorously evaluate the effectiveness of tokens derived from a frozen MAE to solve downstream tasks. This is even without taking into account the extensive breadth of this study on variation in token types, decision rules, and downstream tasks and contexts. Instead, previous work tends to prefer DiNO features for segmentation tasks with works considering frozen MAE features being almost non-existent, possibly due to the known fact that MIM works better when fine-tuned. Yet, a quantitative evaluation of the effectiveness of MAE's features compared to DINO's is currently missing, and our work addresses this gap with a detailed analysis. Our findings suggest that for semantic segmentation, while the downstream performance of MAE's features is inferior to DINO's, in some aspects the gap between them is not as large as one might initially believe. 

Regarding DiNO, methods such as \cite{stego,Lost,tokencut,Amir2021DeepVF,enchancer,cutler} address the unsupervised segmentation task using token feature transformations derived from a frozen backbone. In our work, we differentiate and take a step back to meticulously assess the effectiveness of DiNO's \textbf{vanilla} tokens (without any transformations or extra processing) on downstream tasks using annotations, revealing to some extent the best starting point of those previous approaches. Furthermore, many previous approaches \cite{Lost,stego,tokencut,enchancer,cutler} have applied the cosine similarity rule to the tokens of a frozen DINO backbone, utilizing it as an implicit supervisory signal for semantic similarities. However to our knowledge, a rigorous assessment of its potential is missing from the literature and our work aims to address this, by being the first to assess the effectiveness of DiNO's features with the cosine rule on semantic tasks with ground-truth labels. Our work is also unique in providing a thorough study over the representation power of different token types, being either the attention layer's queries, keys, values, or tokens from either side of the final feed forward transformer block, expanding on the shallow analysis of \cite{dino}. In principle, our findings are aligned with previous work that prefers to use the attention layer's key tokens for semantic segmentation \cite{Lost,tokencut,Amir2021DeepVF,cutler} but also highlights a detailed comparison with the alternative tokens. 
Finally, image classification based on the two SSL approaches is also less explored in the literature \cite{dino,enchancer}, and our work provides a detailed analysis, in terms similar to the segmentation task.

%% file: sections/approach.tex
\section{Approach}
As briefly stated in the preface, our work aspires to address the following questions that innately arise when employing pre-trained ViTs in downstream tasks:
\begin{itemize}
    \item Which self-supervised \textbf{pre-training objective} (MIM, CL, 
    implemented by MAE and DiNO respectively)  produces frozen backbones, 
    which are more aligned to each  \textbf{downstream task} 
    (classification, segmentation)?
    \item Which ViT \textbf{token types} (queries $\q$, keys $\k$, values $\v$ from the final ViT's self-attention block or tokens $\x1, \x2$ from either side of the transformer feed forward block) provide semantically meaningful representations? 
    \item Which \textbf{decision rule} (hyperplane based, cosine similarity based) should be utilized to separate the feature space into semantic regions? 
\end{itemize}
Additionally, we also consider two downstream contexts: standard (where a plethora of labeled examples are available for learning a decision rule) and few-shot (where only a limited number of samples are available for the same purpose). In the following subsections, we aim to clarify these research questions by conducting experiments with combinatorial variability across pre-trained models, tasks, contexts, decision rules, and token types.

\subsection{Self-supervised Pre-Training Objectives} 

This study concentrates on two well-known SSL ViT architectures: MAE \cite{mae} and DiNO \cite{dino}. MAE is part of the group of pre-training techniques focused on masked image modeling, whereas DiNO aligns with self-distillation and contrastive learning approaches. For the sake of computational efficiency, we opted for the smallest pre-trained ViT models accessible to the public (DiNO: ViT-S/8 21M parameters, MAE: ViT-B/16 86M parameters).

\subsection{Downstream Tasks} 

\begin{figure}[h]
    \centering
    \includegraphics[height=5cm, keepaspectratio]{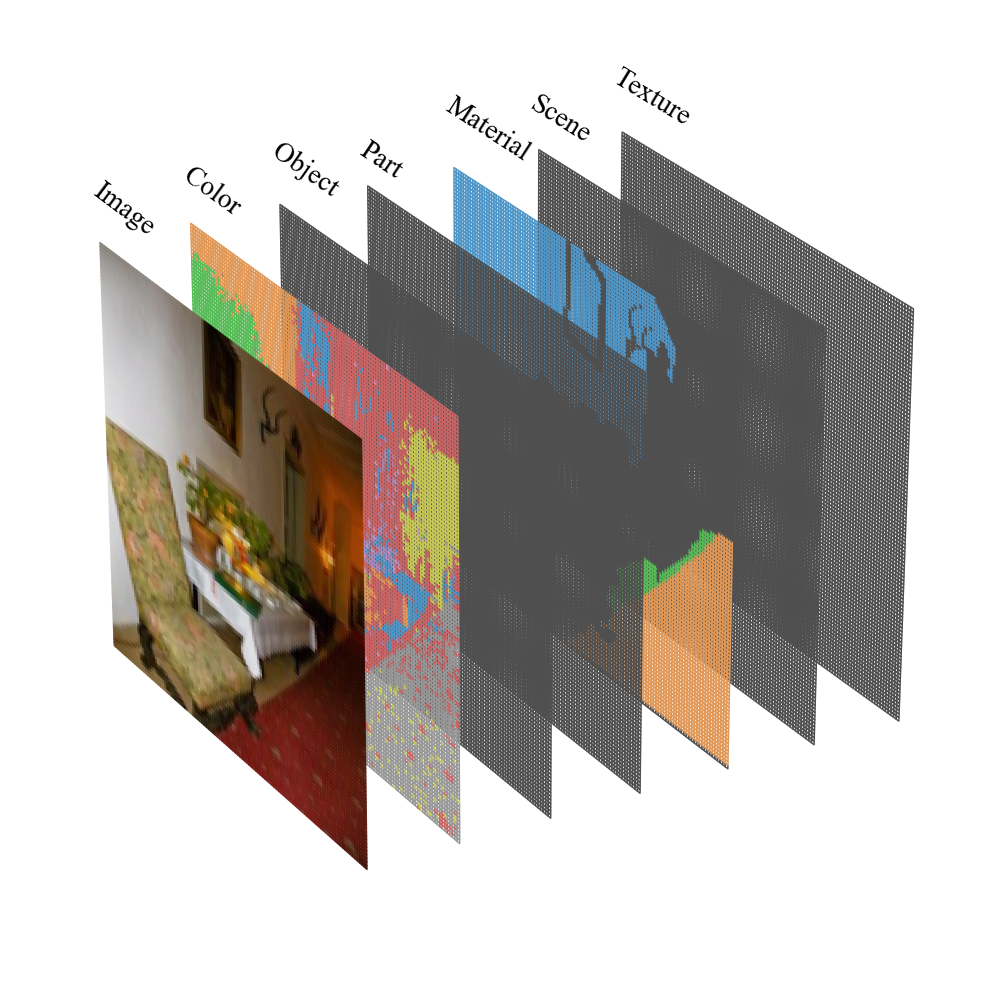}
    \includegraphics[height=5cm, keepaspectratio]{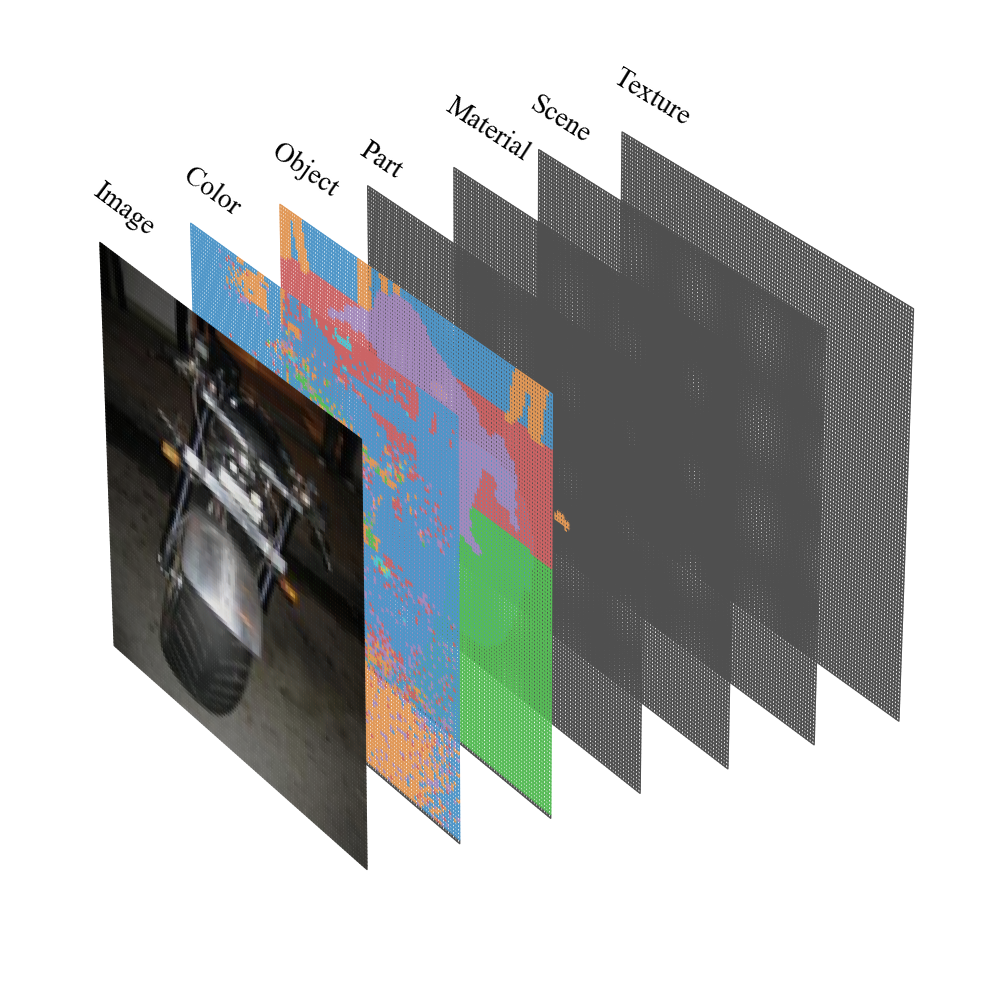}
    \caption{\small Broden samples. Each image in the dataset is associated with multiple segmentation maps, covering six primary categories (color, object, part, material, scene, texture). For instance, the image in the left has a \textit{color} and a \textit{material} category-mapping whereas the image on the right a \textit{color} and an \textit{object} segmentation map.}
    \label{broden_fig}
\end{figure}

We investigate the semantic representation power of ViT tokens in two exemplary downstream tasks: image classification and semantic segmentation. In the context of image classification, we develop a subset of ImageNet \cite{imagenet} resembling ImageNet-Tiny \cite{imagenettiny}, constructed by randomly selecting $550$ samples for each of ImageNet-Tiny's $200$ classes. For image segmentation, we utilize the Broden dataset \cite{netdissect}, which consolidates multiple datasets that are densely annotated \cite{Chen2014DetectWY,Cimpoi2013DescribingTI,Bell2014IntrinsicII,Mottaghi2014TheRO,Zhou2017ScenePT}. Broden encompasses $1197$ concepts distributed across approximately $63K$ images within $5$ distinct concept categories (object, part, material, texture, color). In this research, we have excluded the color category to focus on the remaining categories which are deemed to hold greater semantic significance. Fig.~\ref{broden_fig} demonstrates the extensive annotations present in Broden, which incorporate low-level concept categories, such as material and texture, alongside high-level concepts, such as object and scene.

We address \textbf{both} tasks through a unified binary classification framework, taking inspiration from \cite{Zhou2018InterpretableBD}. Using independent binary classifiers offers a straightforward yet effective learning scheme suited for Broden's multilabel annotation structure. For image classification, we use the [CLS] token as a global feature representation of the entire image, whereas for semantic segmentation, we leverage the corresponding patch tokens to represent individual regions. Consequently, each object --whether the entire image for classification or an image-region for segmentation-- is represented by a single feature vector, which serves as input to a set of binary classifiers. In other words, beyond the typical image classification task, the segmentation task is tackled by treating it as a patch classification problem.

\subsection{Token Types} 
\label{sec:vit-tokens}

\begin{figure}[h!]
    \centering
    \includegraphics[width=0.99\linewidth]{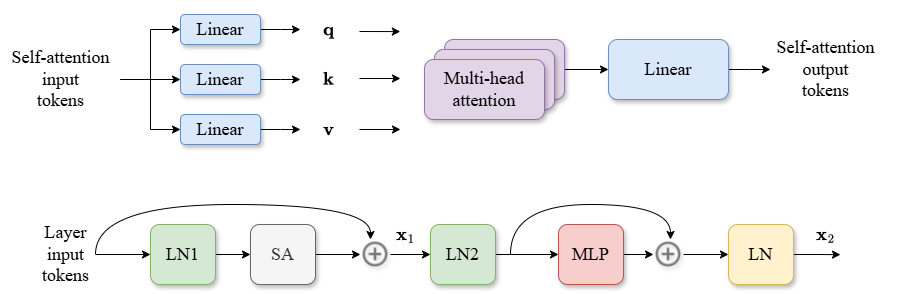}
    \caption{\small (Top) Multi-head attention schematic diagram. $\q, \k, \v \in \mathbb{R}^{D}$ depict the \textit{queries, keys, values} tokens respectively, with $D$ representing the ViT's embedding dimension. (Bottom) Schematic diagram of the transformer's final layer, where LN denotes layer normalization and SA represents the multi-head self-attention mechanism. Note that the final normalization layer (LN) is applied exclusively at the last transformer layer. We denote $\x1 \in \mathbb{R}^{D}$ the transformer tokens prior to the MLP and the second layer normalization layer, while $\x2 \in \mathbb{R}^{D}$ the output-tokens after the MLP (layer output).}
    \label{transformer_layers}
\end{figure}

In our analysis we account for various token types derived from the final transformer layer to address the downstream tasks. We consider the query $\q$, key $\k$, and value $\v$ tokens of the self-attention block (Fig.~\ref{transformer_layers} top), the output of the self-attention block, denoted as $\x1$ and the output tokens of the feed forward block (MLP), referred to as $\x2$ (Fig.~\ref{transformer_layers} bottom).

\subsection{Classifier Decision Rules}

\begin{figure}[h!]
   \hfill
    \includegraphics[height=3.5cm,width=.99\linewidth]{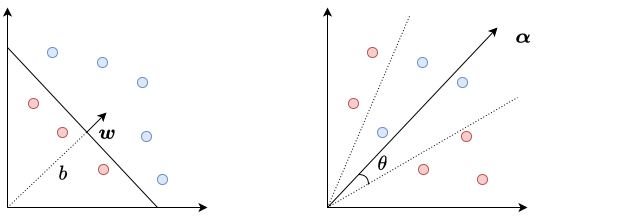}
    \caption{Classifier decision rules. (Left) Hyperplane classifier $(\w, b)$. (Right) Cosine similarity classifier $(\valpha, \theta)$. Each classifier dissects the feature space into two disjoints subspaces. Positively classified samples are depicted in blue, while negatively classified samples are illustrated in red.}
    \label{fig:decision_rules}
\end{figure}

We examine the semantic separability of ViT tokens using two different decision rules: hyperplane-based and cosine similarity-based. As illustrated in Fig.~\ref{fig:decision_rules}, each classification rule is associated with a distinct decision boundary, dissecting the feature space into two disjoint subspaces.

Specifically, the hyperplane rule is comprised of a normal vector $\w$ and a bias term $b$, defining the orientation and position of the hyperplane respectively.  A feature vector $\z$ is classified positively if $\w^T\z - b \geq 0$. In contrast, the cosine decision rule defines a convex cone via a conical axis vector $\valpha$ and an angular threshold $\theta$, such that $\z$ is positively classified if $\arccos(\frac{\z}{\|\vz\|_2} \cdot \frac{\valpha}{\|\valpha\|_2}) \leq \theta$, with $\cdot$ denoting the dot product.

\textbf{Concept Templates:} Both decision rules are associated with a class-specific (in image classification) or concept-specific (in semantic segmentation) \textit{directional} vector, a \textit{threshold} and a \textit{projection function}, which altogether may be utilized to classify a feature vector for the downstream task. We use the term  \textit{concept template} to encompass these attributes and also refrain from making explicit distinction regarding the label type of each downstream task (\textit{image class} vs \textit{patch concept}) as we treat both tasks within a common framework of similar principles. In the rest of the paper we will mostly refer to the downstream task's labels as \textit{concept labels}, when in fact for image classification these labels correspond to image classes.

Formally, given the dimensionality of the embedding space $D$, a feature vector $\z \in \mathbb{R}^D$ and a concept $c \in \mathbb{N}$, the concept template is a triplet $\tau_c \coloneqq (\d, t, f)$, where $\d \in \mathbb{R}^D$ is the directional vector, $t\in \mathbb{R}$ is the threshold and $ f(\z;\d): \mathbb{R}^{D} \rightarrow \mathbb{R} $ is the projection function. 

The concept template $\tau_c$ detects the existence of concept $c$ in the feature vector $\z$ (positive classification) if:
\begin{equation}
    f(\z;\d)   \geq t
\end{equation}
In the case of a hyperplane decision rule: $\d \coloneqq \w, t \coloneqq b$ and $f(\z;\w) \coloneqq \w^T\z$, while for a cosine decision rule: $\d \coloneqq \valpha$, $t \coloneqq \cos(\theta)$ and $f(\z;\valpha) \coloneqq \frac{1}{\|\valpha\|_2 \ \|\z\|_2} \valpha^T \z $. Based on the underlying decision rule, we distinguish two cases of concept templates: \textbf{hyperplane-templates} and \textbf{cosine-templates}.

\subsection{Analysis Framework}
\begin{figure}[t!]
    \centering
    \includegraphics[width=0.98\linewidth]{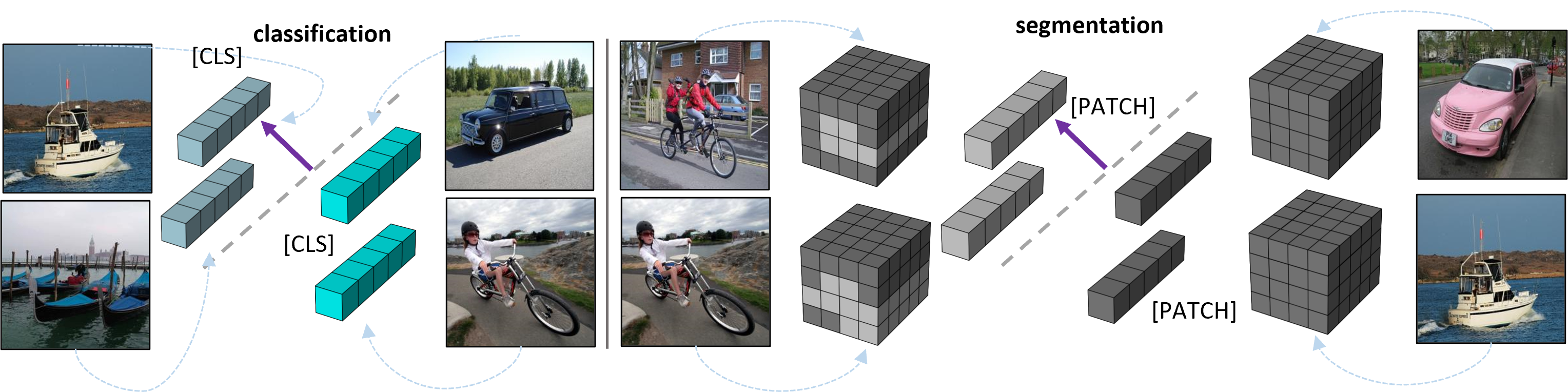}
    \caption{\small Hyperplane decision rule: 
    The class token represents the global image 
    content, while individual image regions are 
    represented by their corresponding patch 
    tokens. A hyperplane is learned for each image class or semantic concept to distinguish positive samples from negative ones.
    }
    \label{fig_t:linear-template}
\end{figure}

\begin{figure}[t!]
    \centering
    \includegraphics[width=0.4\linewidth]{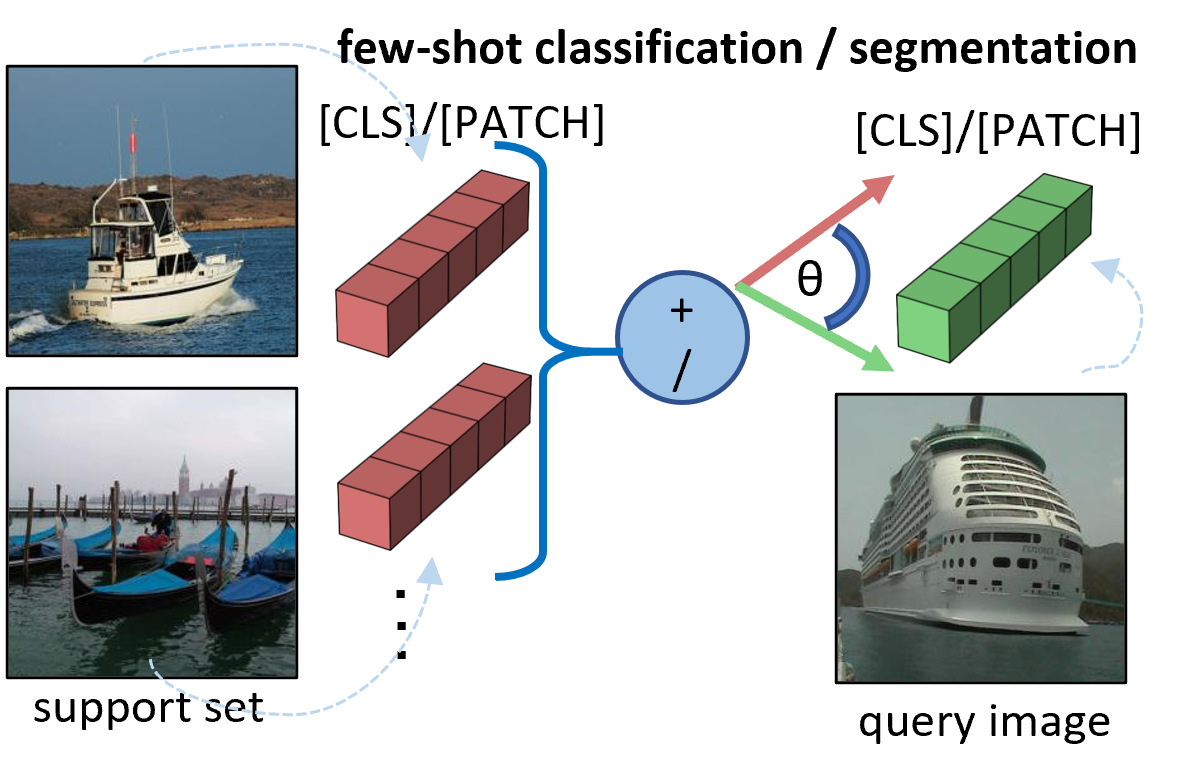}
    \caption{\small Cosine similarity decision rule: 
    In the few-shot context, the concept template's direction is derived by averaging intra-class token representations extracted from the support set. Specifically, a token originating from the query image set is classified positively if the cosine similarity between the token and the template's direction exceeds a threshold $\theta$.}
    \label{fig_t:cosine-template}
\end{figure}

This section provides details on how we learn the concept templates. In this and the rest of the sections, we often use the terms \textit{concept template} and \textit{classifier} interchangeably, preferring the former to emphasize its geometric interpretation and the latter to focus on its functional application.

\subsubsection{Hyperplane Templates:}
To compute hyperplane templates, for each concept, we learn a hyperplane classifier ($\w,b$) with the process illustrated in Fig.~\ref{fig_t:linear-template}. Given a training feature dataset $D_f: \{(\z_i, c_i), 
i = 1, \dots, N\}$, where $\z_i \in \mathbb{R}^D$ represents the feature vector of an object (image/image-region) and $c_i \in \mathbb{N}$ represents its ground-truth label, we construct a positive sample pool for each concept $c$, denoted as $D^+_c = 
\{\z_i \ |\  (\z_i, c_i) \in D_f, \ c_i = c\}$, and a corresponding negative sample pool, $D^-_c = \{\z_i \ | \ (\z_i, c_i)
\in D_f, \  c_i \neq c\}$, where $|D^-_c| \gg |D^+_c|$. In semantic segmentation, when 
forming the negative sample pool for a concept $c_i$, we only consider concepts within 
the same primary category as $c_i$. To manage the significant class imbalance between 
the two sample pools, we initially limit the size ratio of $D^-_c:D^+_c$ to be no more 
than $20:1$ by random subsampling. During template learning, we conduct five rounds of 
hard negative mining, following \cite{Zhou2018InterpretableBD}. In each of these 
rounds, the hyperplane template is fitted to the mined dataset over 3 epochs, ensuring 
a positive-to-negative sample ratio of $1:2$. The evaluation of \textbf{each} learned 
template is performed on a reserved test-set (approximately $10K$ image samples for ImageNet and $18K$ image samples for Broden from its validation split) via a set of \textbf{balanced} binary 
classification metrics.

\subsubsection{Cosine Templates:} Since the cosine decision rule is frequently utilized in unsupervised settings \cite{Lost,stego,tokencut,cutler,enchancer} including few-shot contexts, we \textbf{explicitly} consider learning cosine-templates in a few-shot regime by constructing support-query image sets for template learning and evaluation. The directional vector $\valpha$ and similarity threshold $t \doteq \text{cos}(\theta)$ of cosine templates, are computed in a non-parametric 1-way-k-shot setting. For a concept $c \in \mathbb{N}$, we construct a \textit{support image set} 
$S_{c,k}$ by randomly sampling $k$ training images that contain $c$. 
$S_{c, k}$ is further processed to construct the respective positive and negative \textit{support feature pools} $D_{c,k}^+, D_{c,k}^-$; Notice that for image classification $D_{c,k}^- = \emptyset$, as every image in the support set is mapped to a single feature vector ([CLS] token). The cosine template's directional vector $\valpha$ is then computed by averaging the positive support features: 
\begin{equation}
    \valpha = \frac{1}{|D_{c, k}^+|} \sum_{\z \in D_{c,k}^+}\z
\end{equation}
while the angular threshold $\theta$ is computed by maximizing the F1-score of the classifier $\tau_c$ on the \textit{support feature set}:
\begin{equation}
    \label{eq:cosine-theta}
    \theta = \underset{\hat{\theta}}{\text{argmax}}(F1(\hat{\theta}, D_{c,k} ; \valpha, f))
\end{equation}
where $F1$ is the F1 score of a classifier $\tau_c = (\valpha, t, f)$ computed on the support feature set $D_{c,k} = D_{c,k}^+ \cup 
D_{c,k}^- $, given the directional vector $\valpha$ and the cosine-similarity projection function as $f$. Due to the fact that we use an empty $D_{c,k}^-$ for image classification, in Eq. (\ref{eq:cosine-theta}) we consider the smallest possible angle $\theta$ that maximizes F1 score. The overall process is illustrated in Fig. \ref{fig_t:cosine-template}. Furthermore, we vary $k \in \{1, 5, 10, 50, 100, 500\}$, leveraging different proportions of the available data. Finally, the templates are evaluated on a balanced randomly sampled query test set of $50$ positive and $50$ negative images using the same set of balanced binary classification metrics as in the hyperplane templates. Due to the stochastic nature of this 1-way-k-shot setting, we average and present the results from $N=10$ independent trials reporting mean scores and their standard deviation.

%% file: sections/results.tex
\section{Experimental Results}
The subsequent subsections detail the outcomes of our comprehensive experimental evaluation, structured by downstream task and decision rule. In our analysis, the term \textit{token performance} is used to denote the efficacy of concept templates that incorporate a particular token. It is important to highlight that for image classification tasks, the mentioned tokens refer to the [CLS] tokens, while for image segmentation, they pertain to patch tokens. Lastly, we underline that all the binary performance metrics presented in this work are \textbf{balanced}.

\subsection{Task: Classification. Rule: Hyperplane}

\textbf{TLDR:} We observe a substantial disparity in the classification performance of the hyperplane template between the pre-trained MAE and DiNO models. While MAE tokens resemble the performance of random classifiers, DiNO demonstrates exceptional classification capacity. Specifically, DiNO's $\x2$ token is particularly well-suited for classification tasks via linear probing, while MAE should not be considered in this context.

\begin{figure}[h]
    \centering
    \includegraphics[width=0.49\linewidth]{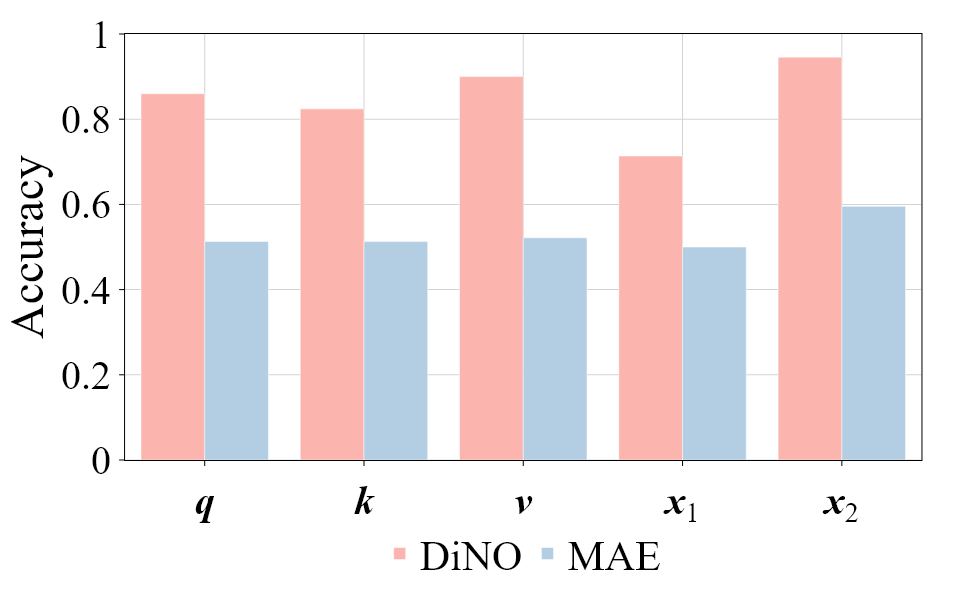}
    \includegraphics[width=.49\linewidth,]{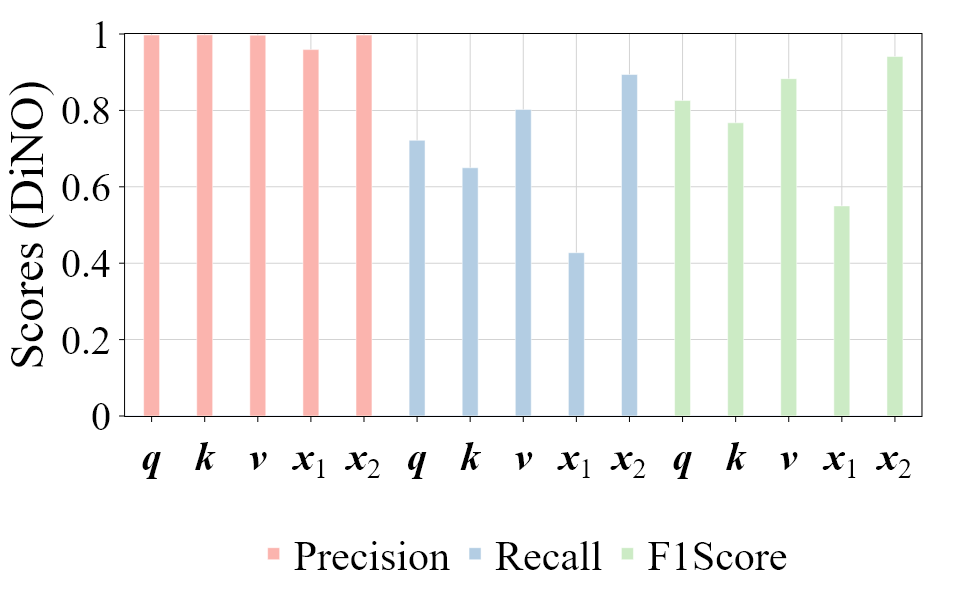}
     \caption{Hyperplane-template classification: (Left) Accuracy between DiNO and MAE tokens. (Right) Precision, recall and F1 metrics for DiNO tokens.}
     \label{fig:ibd_classif_results}
\end{figure}
\textbf{Details}: Fig.~\ref{fig:ibd_classif_results} (Left) compares MAE and DiNO tokens in terms of accuracy. Most of MAE tokens approximately score an accuracy of 0.5, which is equivalent to a random classifier. This may be attributed to the fact that the [CLS] token is not participating in the MAE's loss function. In contrast, DiNO attains its maximum accuracy with $\x2$ (0.946).  
A detailed analysis of DiNO's token performance is presented in Fig.~\ref{fig:ibd_classif_results} (Right). We observe near-perfect precision for $\q$, $\k$, $\v$, and $\x2$ (> 0.99), while $\x1$ achieves a precision of 0.96. This enables the construction of a hyperplane with minimal false positives (FP) across all tokens. Furthermore, $\x2$ exhibits the highest recall (0.89), followed by $\v$ (0.80), $\q$ (0.72) and $\k$ (0.65). These results indicate that $\x2$ provides the optimal linear separability of semantic concepts.

Notably, $\x1$ demonstrates the lowest performance across all evaluated metrics. To better understand this phenomenon, we also assess the performance of $\x1$ after 
layer normalization, which we denote as $\vx_{n}$. Table~\ref{tab:dino_metrics} presents the impact of the normalization layer on DiNO's $\x1$ hyperplane classification metrics. Layer normalization positively affects  the semantic linear separability of the feature space. However, a more detailed analysis of the effects of layer normalization is beyond the scope of this work.

\begin{table}[h]
    \centering
    \caption{Layer normalization effects on DiNO's $\x1$ performance metrics.}
    \begin{tabular}{lcccc}
        \toprule
        \textbf{DiNO} & \textbf{Accuracy} & \textbf{Precision} & \textbf{Recall} & \textbf{F1} \\
        \midrule
        $\x1$ & 0.714 & 0.959& 0.427 & 0.550 \\
        $\vx_n$ & 0.940 & \textbf{0.997} & 0.884 & 0.935 \\
        $\x2$ & \textbf{0.946}  & \textbf{0.997} & \textbf{0.894} & \textbf{0.941} \\
        \bottomrule
    \end{tabular}
    \label{tab:dino_metrics}
\end{table}

\subsection{Task: Classification. Rule: Cosine}
\textbf{TLDR}: Similar to hyperplane-based classification, DINO 
outperforms MAE under the cosine similarity decision rule. 
Notably, DINO's $\x1$ token achieves the 
highest accuracy and F1 scores. Furthermore, MAE shows 
substantial improvement with cosine templates compared to 
the hyperplane decision rule, with its $\k$ 
token yielding the highest accuracy and F1 score in this 
context. Finally, increasing the support set size beyond 50 
samples results in diminishing gains in average accuracy 
and F1 scores for both models.

\begin{figure}[h]
    \centering
    \includegraphics[width=0.49\linewidth]{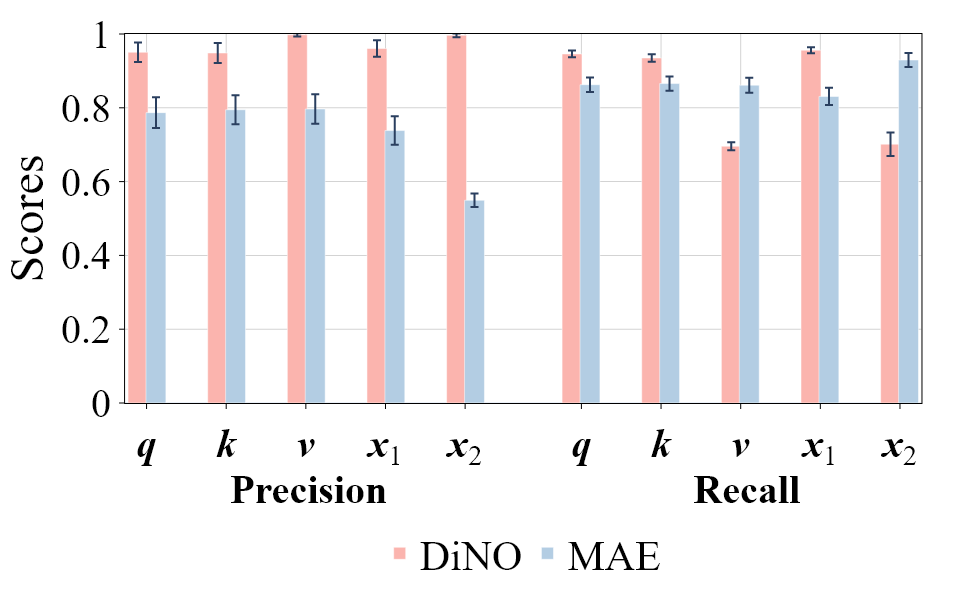}
    \includegraphics[width=.49\linewidth,]{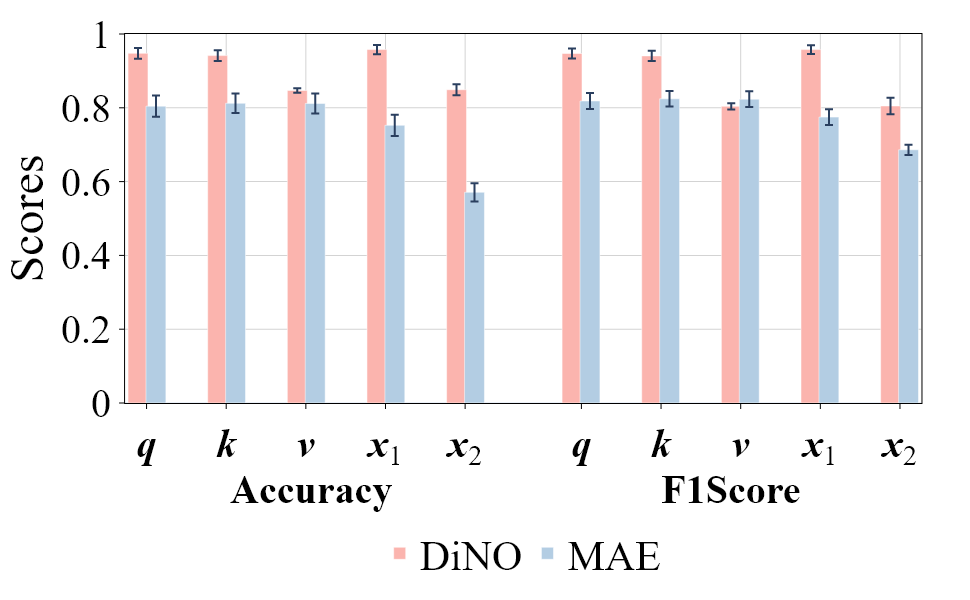}
     \caption{Cosine-template classification with $k = 500$ support samples per concept.  (Left)  Precision and recall comparison between MAE and DiNO tokens. (Right)  Accuracy and F1 score comparison. The error bars denote the standard deviation across $10$ independent trials.}
     \label{fig:fste_classif}
\end{figure}

\textbf{Details}: Fig.~\ref{fig:fste_classif} shows the classification metrics for DiNO and MAE tokens using the cosine decision rule, averaged over 10 independent trials with $k=500$ support images per concept. DiNO's $\x1$ emerges as the optimal token, achieving the highest accuracy (0.958 ± 0.01) and F1 score (0.958 ± 0.01), while $\q$ and $\k$ perform similarly. Although all DiNO tokens demonstrate high precision, $\v$ and $\x2$ exhibit the lowest recall in this setting.
For MAE, $\k$ achieves the highest accuracy (0.812 ± 0.03) and F1 score (0.824 ± 0.02), while $\q$ and $\v$ demonstrate similar performance. Notably, $\x2$ exhibits the highest recall (0.929 ± 0.02), making it particularly well-suited for critical risk detection applications where minimizing false negatives (FN) is essential.  
\par 

\begin{figure}
    \centering
    \includegraphics[width=0.49\linewidth]{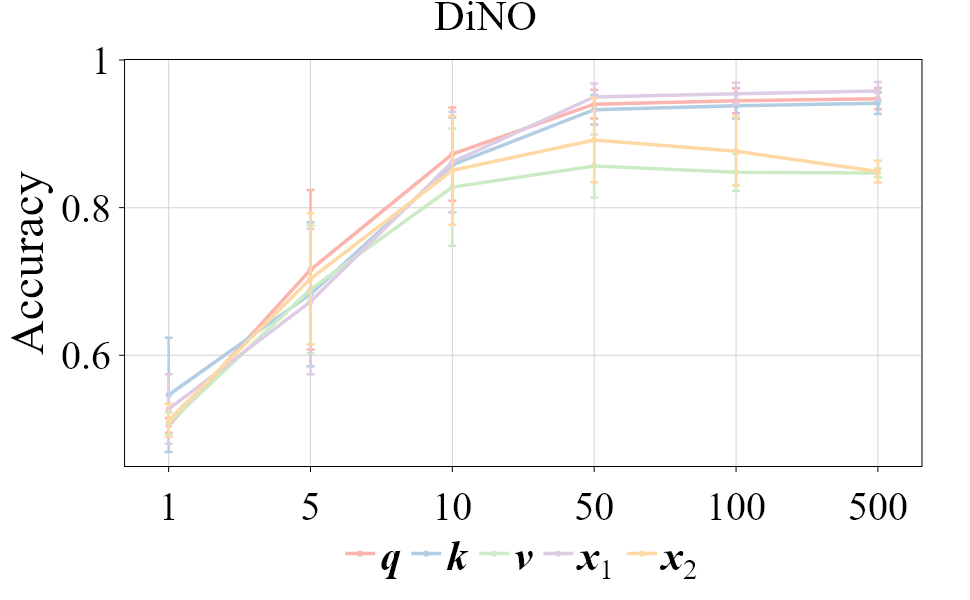}
    \includegraphics[width=0.49\linewidth]{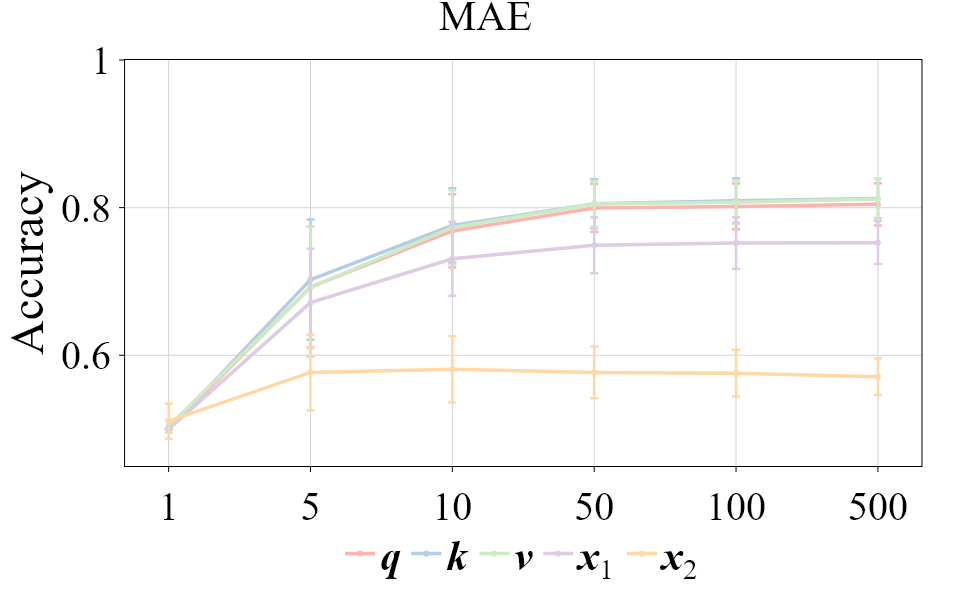}
    \caption{Cosine-template classification accuracy for  $k \in \{1, 5, 10, 50, 100, 500\}$ support samples per concept, for DiNO (Left) and MAE (Right).  Error bars denote standard deviation across 10 independent trials.} 
    \label{fig:fste_k_accuracy}
\end{figure}

Fig.~\ref{fig:fste_k_accuracy} illustrates the impact of  $k$ (number of support samples used to compute cosine-templates) on model accuracy. Notably, performance gains diminish significantly beyond $50$ samples. However, increasing the number of support samples leads to a more representative support set, thereby reducing the standard deviation across trials. 

\subsection{Task: Segmentation. Rule: Hyperplane}
\textbf{TLDR}: Both MAE and DINO demonstrate strong and comparable hyperplane-template accuracy, yet inferior to the scores for image classification. Between the two pre-trained models, DINO achieves a higher overall F1 score. Notably, $\k$ is the optimal token in terms of overall accuracy and F1 score for both models. However, while $\k$ consistently yields the highest F1 score across all concept categories in DINO, MAE shows a slight advantage for $\x2$ over $\k$ when considering textures, objects, or scenes.

\begin{figure}[h!]
    \centering
    \includegraphics[width=0.49\linewidth]{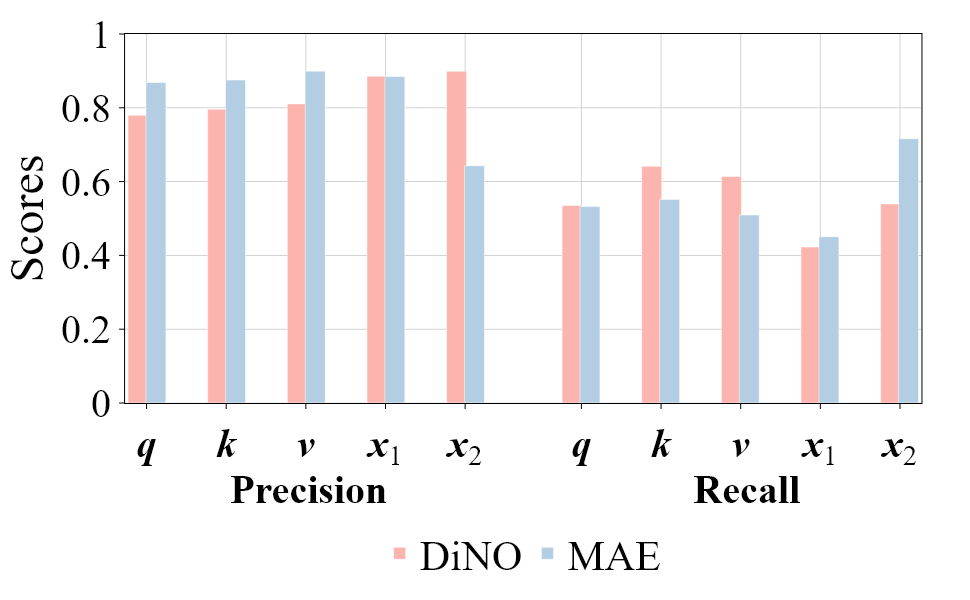}
    \includegraphics[width=0.49\linewidth]{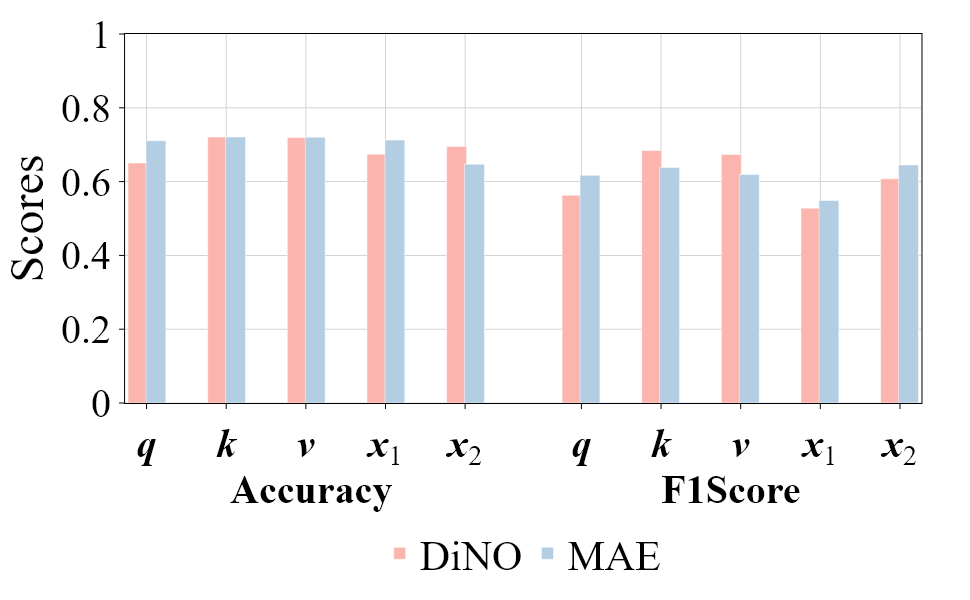}
    \caption{Hyperplane-template segmentation: (Left)  Precision and recall comparison between MAE and DiNO 
    tokens. 
    (Right) Accuracy and F1 score comparison between MAE and DiNO tokens.}
    \label{fig:ibd_seg_overall}
\end{figure}

\textbf{Details}: Fig.~\ref{fig:ibd_seg_overall} presents the overall hyperplane-template segmentation performance of DiNO and MAE tokens. 
Among DiNO tokens, $\k$ achieves the highest accuracy (0.721) and F1 score  (0.684), while $\v$ 
attains similar accuracy (-0.001) but a slightly lower F1 score (-0.01).
DiNO's $\x2$ exhibits the highest precision (0.899) making it particularly well-suited for quality assurance applications where minimizing false positives (FP) is essential.
MAE's $\k$ achieves the highest accuracy (0.721), while $\x2$ attains the 
highest F1 score (0.645). Comparing the two, $\k$ appears to be the optimal choice, 
with a significantly higher accuracy (+0.07) and only a slight reduction in 
F1 score (-0.01). On the other hand, $\v$ demonstrates the highest precision 
(0.899), while $\x2$ excels in recall (0.716). Notably, $\x2$ shows a 
substantial precision drop compared to $\x1$ (-0.24), coupled with a 
significant recall increase (+0.27). This suggests that critical semantic information may be lost in $\x2$, likely in favor of low-level textural patterns, as $\x2$ tokens are processed through a decoder for masked patch reconstruction.
\begin{figure}
    \centering
    \includegraphics[width=0.49\linewidth]{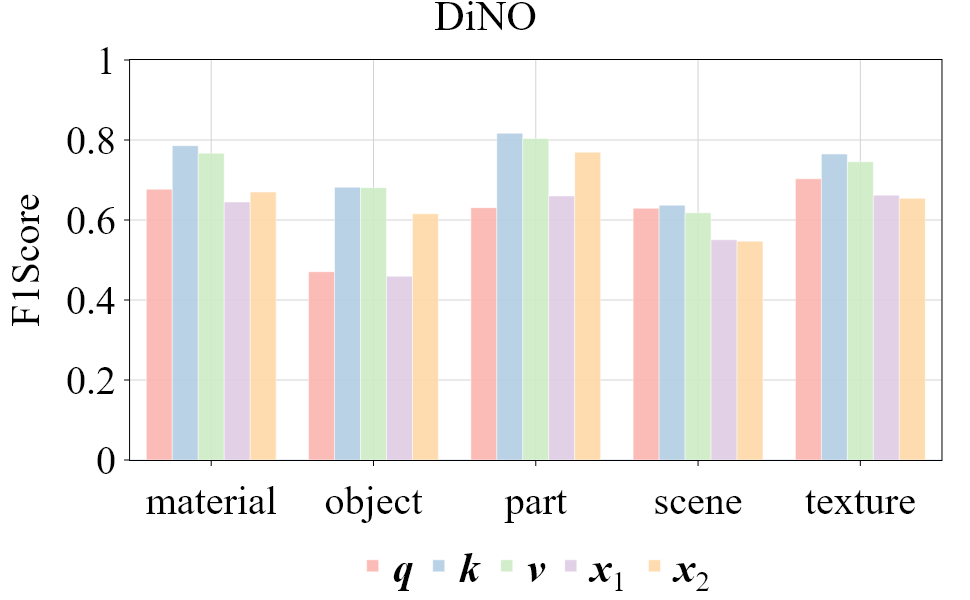}
    \includegraphics[width=0.49\linewidth]{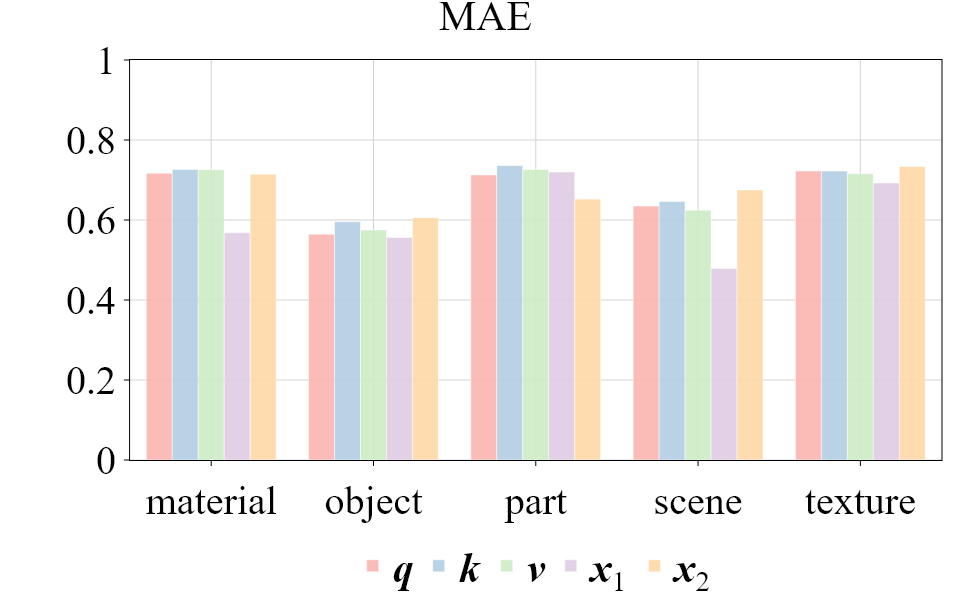}
    \caption{F1 score for DiNO (Left) and MAE (Right) templates, grouped by label category. 
    The scores for each concept template, are grouped and averaged according to their Broden primary semantic category (material, object, part, scene, texture).}
    \label{fig:ibd_seg_cat_f1score}
\end{figure}

Fig.~\ref{fig:ibd_seg_cat_f1score} presents the F1 scores of hyperplane templates, grouped by  semantic category. DiNO’s $\k$ token consistently outperforms others regardless of the semantic category. Among all concept categories, DiNO performs better in part (0.817), material (0.786), and texture (0.765) but is less effective in object (0.682) and scene (0.637) categories. This pattern suggests that DiNO’s $\k$ token excels at segmenting fine-grained semantic concepts, aligning with prior findings \cite{Amir2021DeepVF}.
For MAE, the $\k$ token achieves the highest F1 scores in part (0.734) and material (0.727), whereas $\x2$ leads in object (0.606), scene (0.676), and texture (0.734). Notably, the most significant disparity occurs in the part category, where $\k$ significantly outperforms $\x2$ (+0.08). Interestingly DiNO achieves higher F1 score compared to MAE, in all categories except for scene (-0.13).

\begin{figure}[h!]
    \centering
    \includegraphics[width=0.49\linewidth]{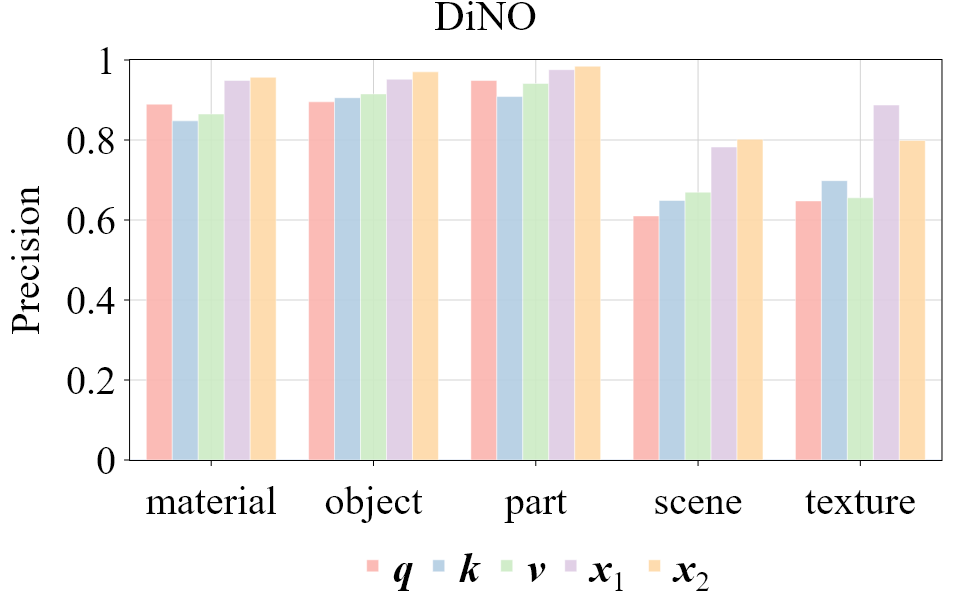}
    \includegraphics[width=0.49\linewidth]{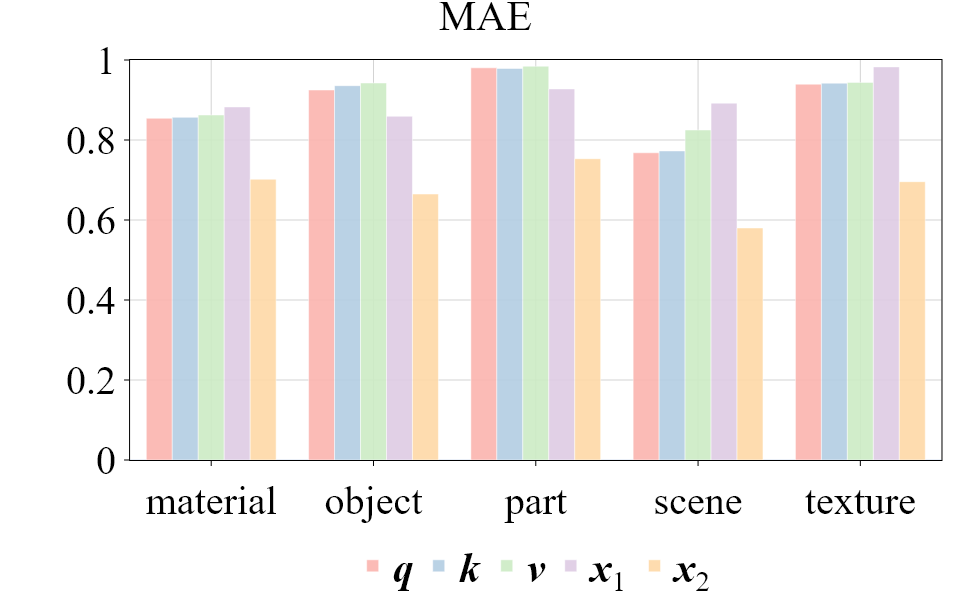}
    \includegraphics[width=0.49\linewidth]{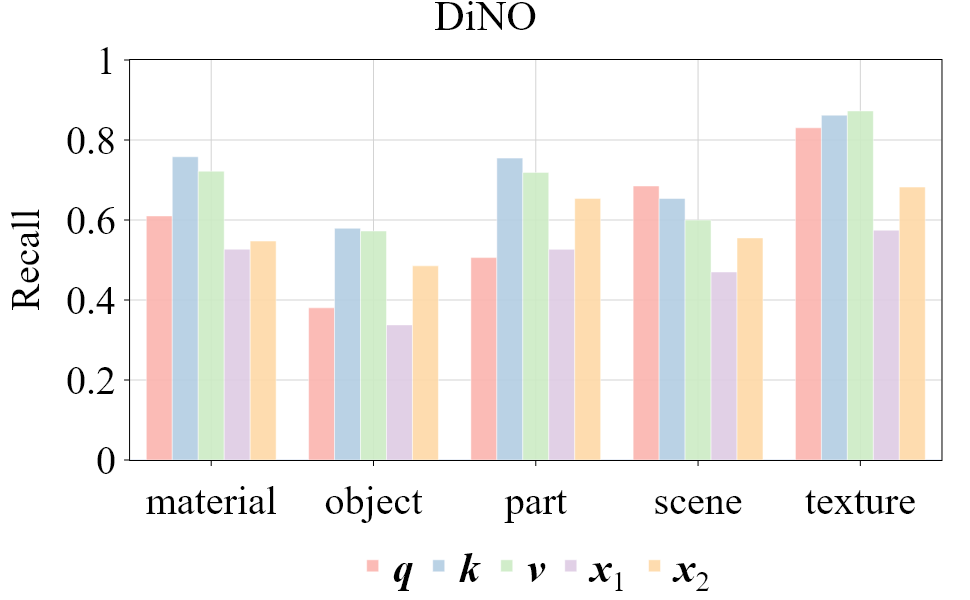}
    \includegraphics[width=0.49\linewidth]{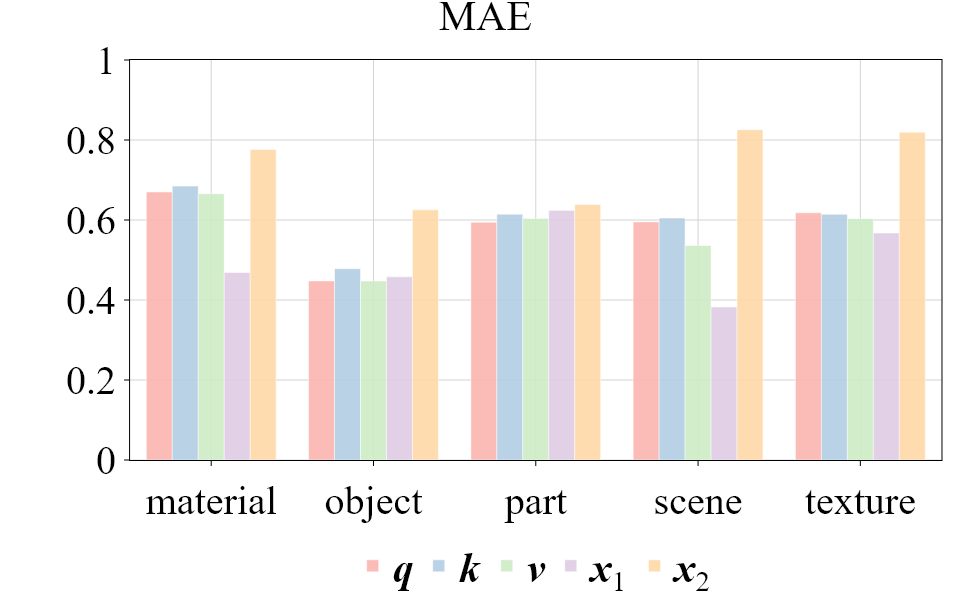}
    \caption{Precision (Top) and recall (Bottom) for DiNO (Left) and MAE (Right) tokens, grouped by label category.}
    \label{fig:ibd_seg_prec_recall}
\end{figure}

Fig.~\ref{fig:ibd_seg_prec_recall} further examines precision and recall across concept categories. In terms of precision, DiNO’s $\x2$ token achieves the highest overall score (Fig.~\ref{fig:ibd_seg_overall}), a trend that persists across most categories, except for texture, where $\x1$ exhibits superior precision (+0.09). For MAE, $\v$ achieves the highest precision in object and part categories, while $\x1$ is the most precise in material, scene, and texture. Notably, MAE’s $\x1$ consistently outperforms $\x2$ in precision across all Broden categories. When analyzing recall, DiNO’s $\k$ token demonstrates the best performance in material, object, and part categories, whereas $\q$ and $\v$ emerge as the top-performing tokens for scene and texture, respectively. Regarding recall for MAE, $\x2$ consistently performs best across all categories.

Cross-model comparisons reveal that MAE’s $\v$ or $\x1$ tokens achieve higher precision than DiNO in part, scene, and texture categories, while DiNO tokens exhibit superior precision in material and object categories, reinforcing its strength in segmenting individual structures. 

\subsection{Task: Segmentation. Rule: Cosine}
\textbf{TLDR}: For both MAE and DiNO, the utilization of the cosine-decision rule is 
evidently inferior to hyperplane-templates, as their overall accuracy 
across all concepts is not significantly superior to a random-classifier 
($\approx 0.6$). However,
both models can achieve notable accuracy  and F1 scores for 
textural concepts.

\begin{figure}[!h]
    \centering
    \includegraphics[width=.49\linewidth]{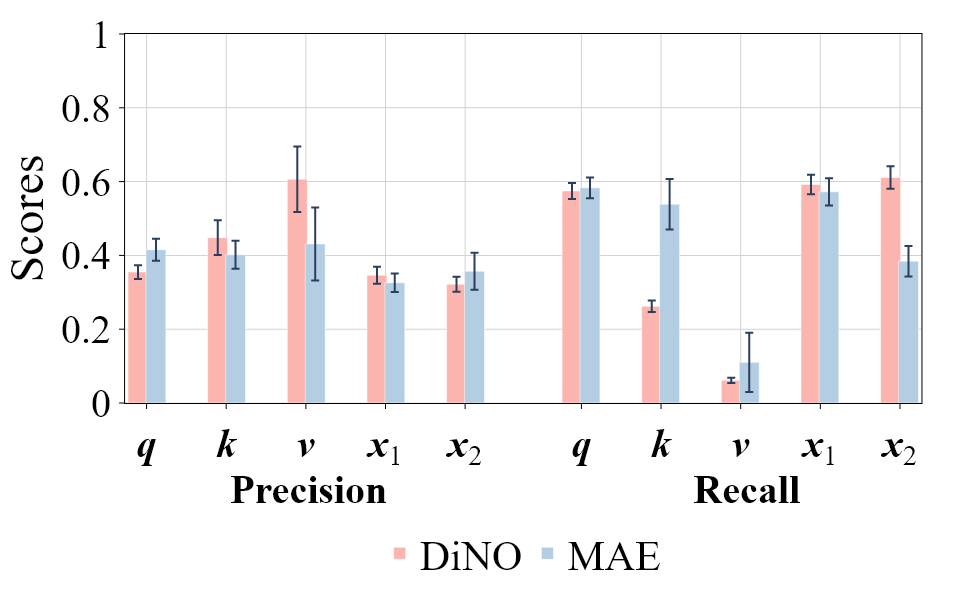}
    \includegraphics[width=.49\linewidth]{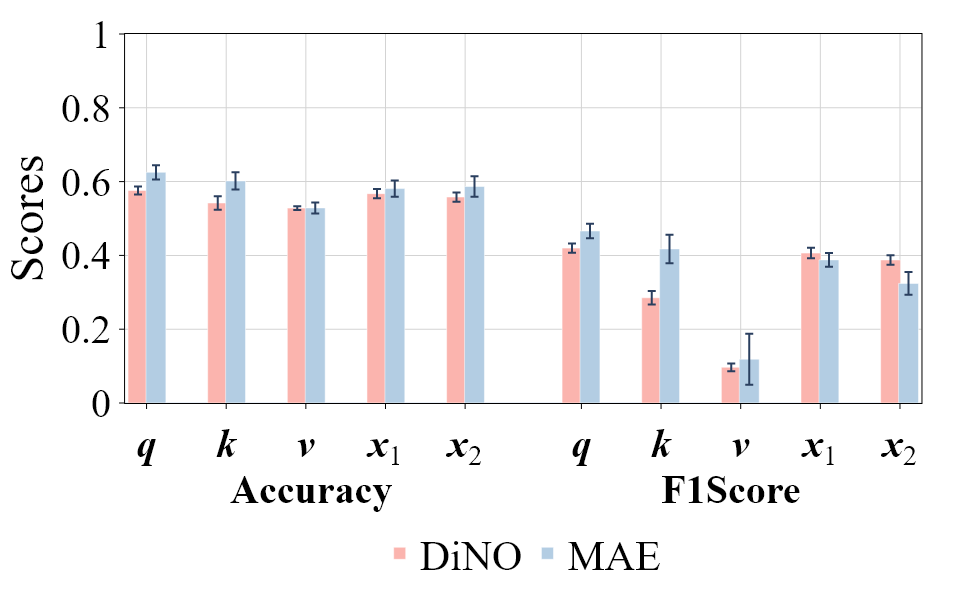}
    \caption{Cosine-template segmentation with $k = 500$ support samples per concept. (Left)
 Precision and recall comparison between MAE and DiNO tokens. (Right) Accuracy and F1 score comparison. The error bars denote the standard deviation
across 10 independent trials.}
    \label{fig:fste_seg_overall}
\end{figure}

\textbf{Details}: Fig.~\ref{fig:fste_seg_overall} presents the overall segmentation metrics for DiNO and MAE tokens under the cosine decision rule, averaged over 10 trials with $k = 500$ support images per concept. In both models, $\q$ tokens achieve the highest accuracy (DiNO: 0.574, MAE: 0.622) and F1 scores (DiNO: 0.419, MAE: 0.464). While MAE outperforms DiNO, both models perform significantly worse compared to the hyperplane decision rule, highlighting the limitations of the cosine decision rule in this setting. 

\begin{figure}[!h]
    \centering
    \includegraphics[width=0.49\linewidth]{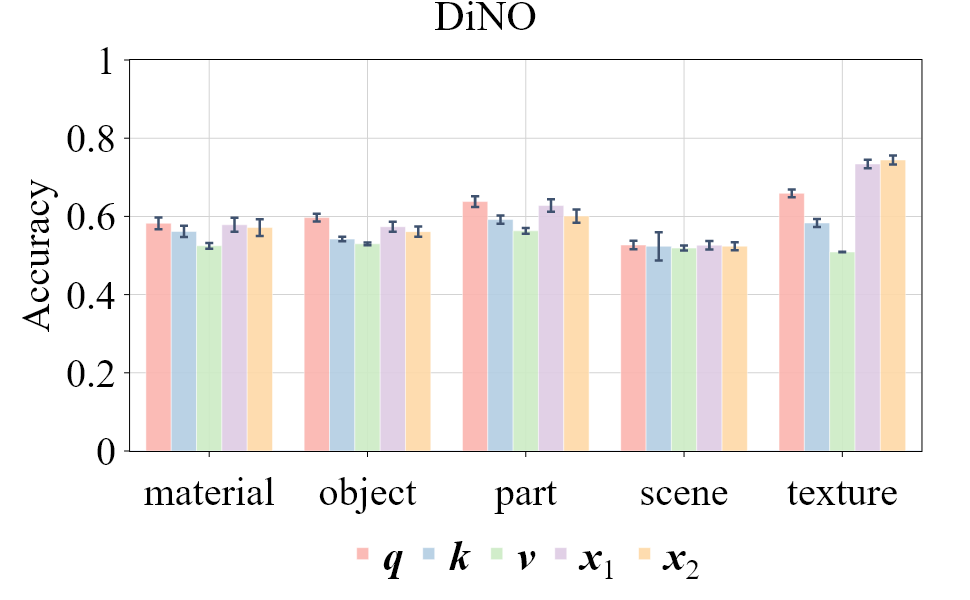}
    \includegraphics[width=0.49\linewidth]{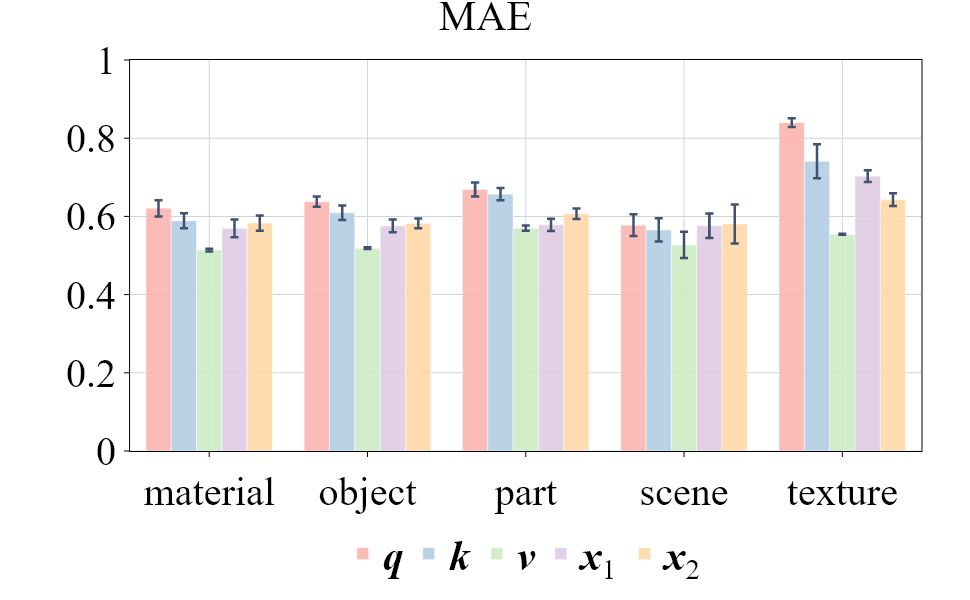}
    \includegraphics[width=0.49\linewidth]{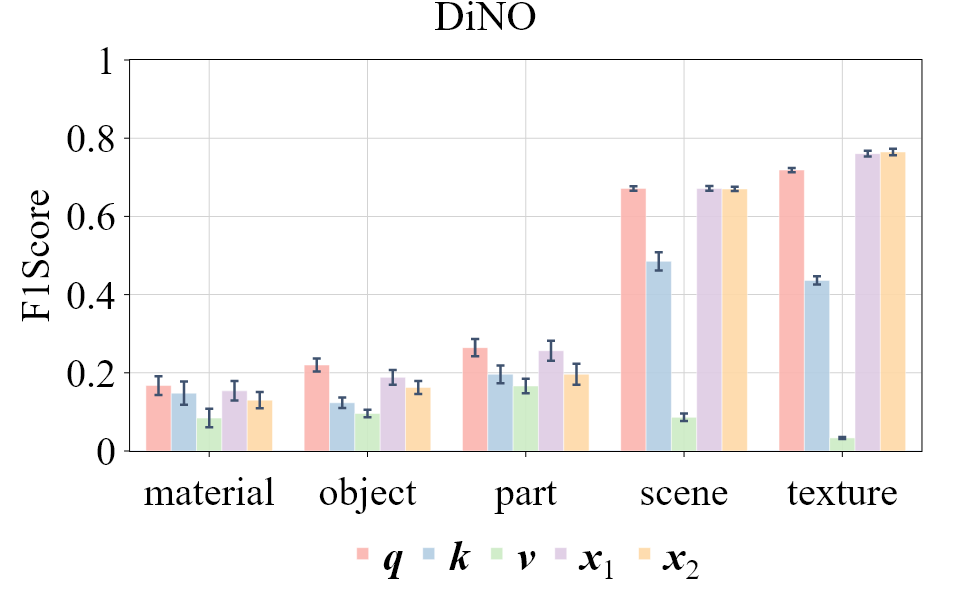}
    \includegraphics[width=0.49\linewidth]{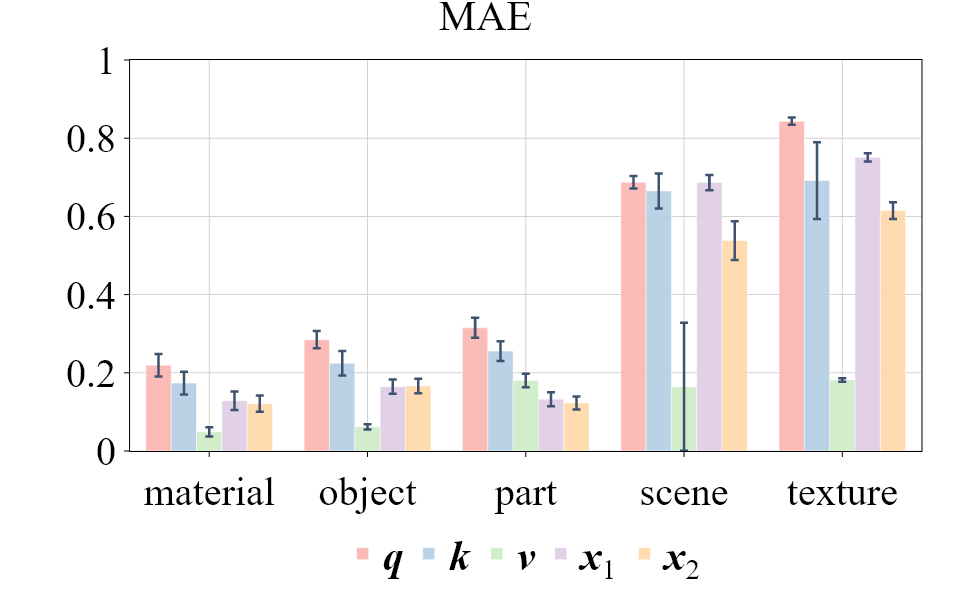}
    \caption{Cosine-template segmentation with $k = 500$ support samples per concept. (Top)  Accuracy for DiNO (Left) and MAE (Right) tokens, grouped by label category. (Bottom) F1 score for DiNO (Left) and MAE (Right) tokens, grouped by label category. The error bars denote the standard deviation across 10 independent trials.}
    \label{fig:fste_seg_cat}
\end{figure}

\begin{figure}[!h]
    \centering
    \includegraphics[width=0.47\linewidth]{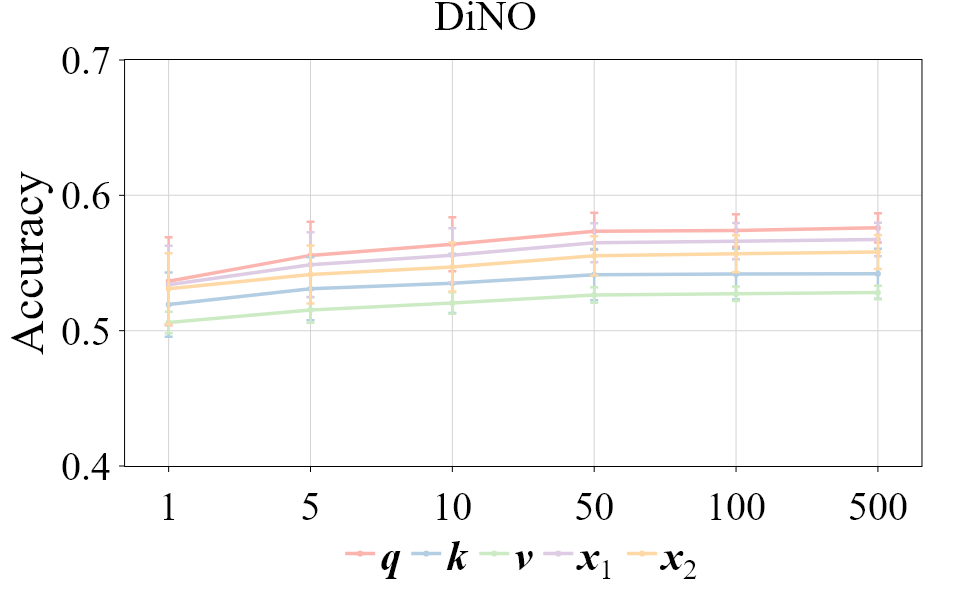}
    \includegraphics[width=0.47\linewidth]{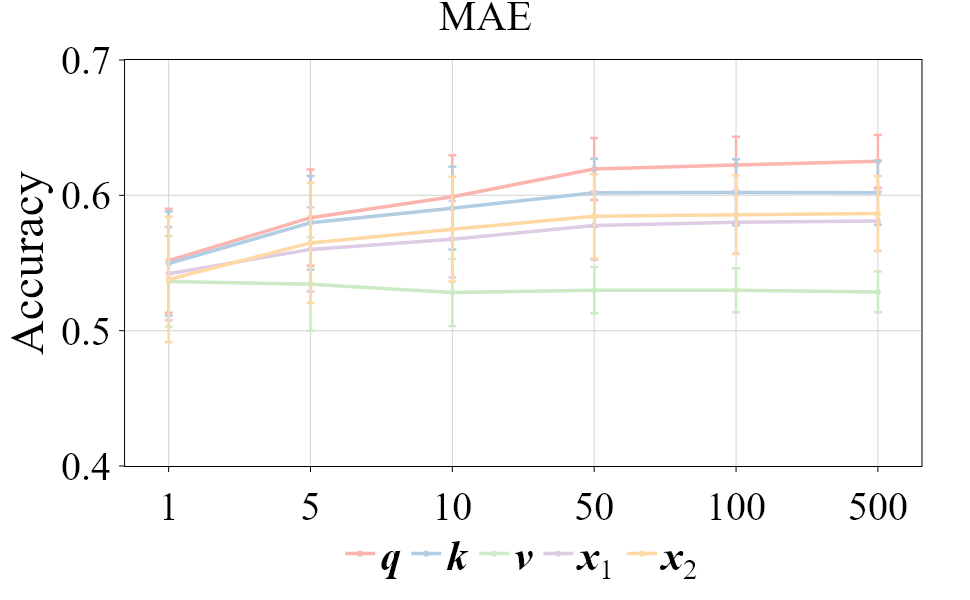}
    \caption{Cosine-template segmentation accuracy for $k \in \{1, 5, 10, 50, 100, 500\}$ support samples per concept, for DiNO (Left), MAE (Right). Error bars denote standard deviation across 10 independent trials.}
    \label{fig:fste_seg_k}
\end{figure}

Figure~\ref{fig:fste_seg_cat} shows the accuracy and F1 scores for cosine templates, grouped by semantic category. Notably, both models perform well on textural concepts, and partially well (low accuracy, but higher F1 score) on scenes. MAE’s $\q$ token achieves an average accuracy of 0.840 and an F1 score of 0.844, while DiNO’s $\x2$ token reaches an average accuracy of 0.744 and an F1 score of 0.765. Fig.~\ref{fig:fste_seg_k} illustrates the impact of $k$ (number of support samples used to compute cosine-templates) on model accuracy. Similar to cosine-template classification, performance gains diminish significantly beyond $50$ samples. 

\subsection{Qualitative Results} 
In the following subsection, we qualitatively examine the segmentation capabilities of learned concept templates on \textbf{unseen} image samples. Based on our previous analysis, we use the $\k$ tokens for hyperplane templates and the $\q$ tokens for cosine templates for both DiNO and MAE.

\begin{figure}[h!]
    \centering
    \begin{subfigure}[t]{0.49\linewidth}
        \includegraphics[width=\linewidth]{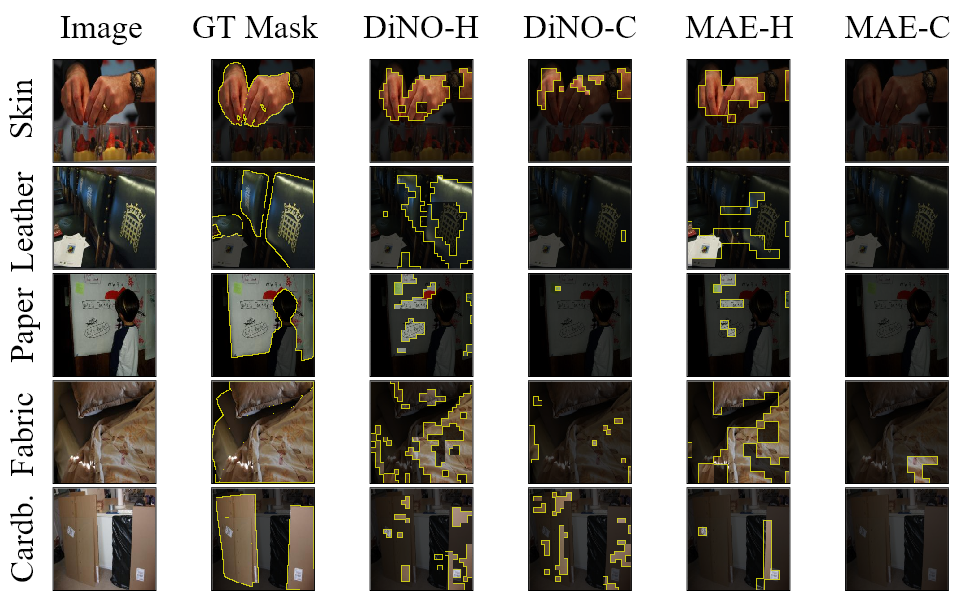}
        \caption{{\scriptsize Material: Skin, leather, paper, fabric, cardboard.}}
        \label{fig:sub1}
    \end{subfigure}
    \hfill
    \begin{subfigure}[t]{0.49\linewidth}
        \includegraphics[width=\linewidth]{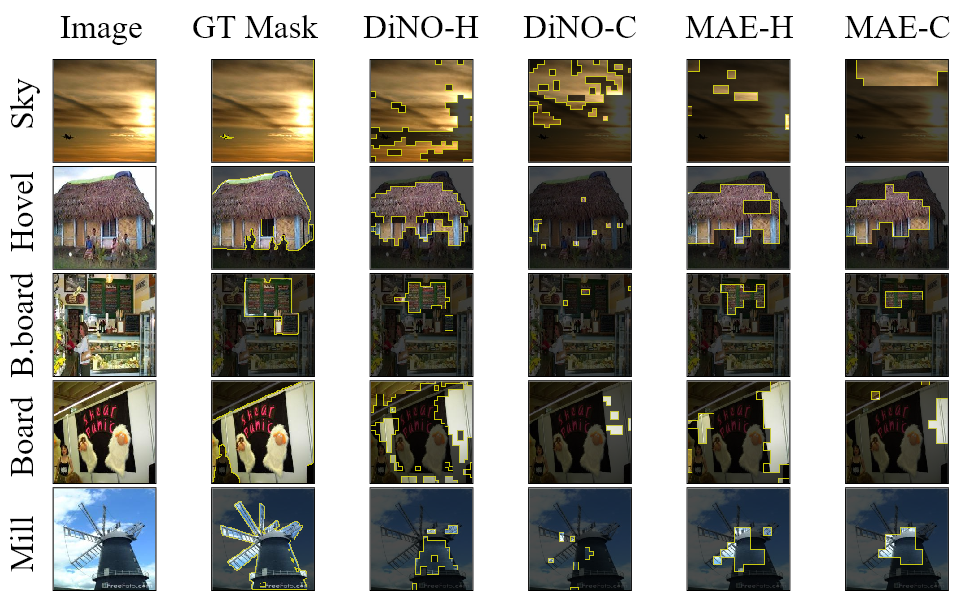}
        \caption{{\scriptsize Object: Sky, hovel, bulletin-board, board, windmill.}}
        \label{fig:sub2}
    \end{subfigure}
    
    \vspace{0.5cm} %
    
    \begin{subfigure}[t]{0.49\linewidth}
        \includegraphics[width=\linewidth]{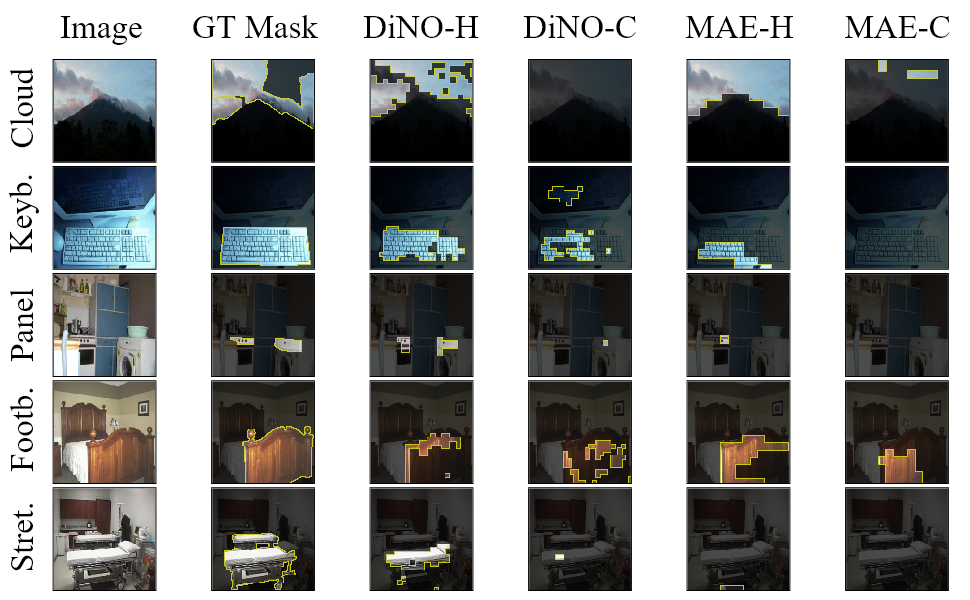}
        \caption{{\scriptsize Part: Cloud, keyboard, button-panel, footboard, stretcher.}}
        \label{fig:sub3}
    \end{subfigure}
    \hfill
    \begin{subfigure}[t]{0.49\linewidth}
        \includegraphics[width=\linewidth]{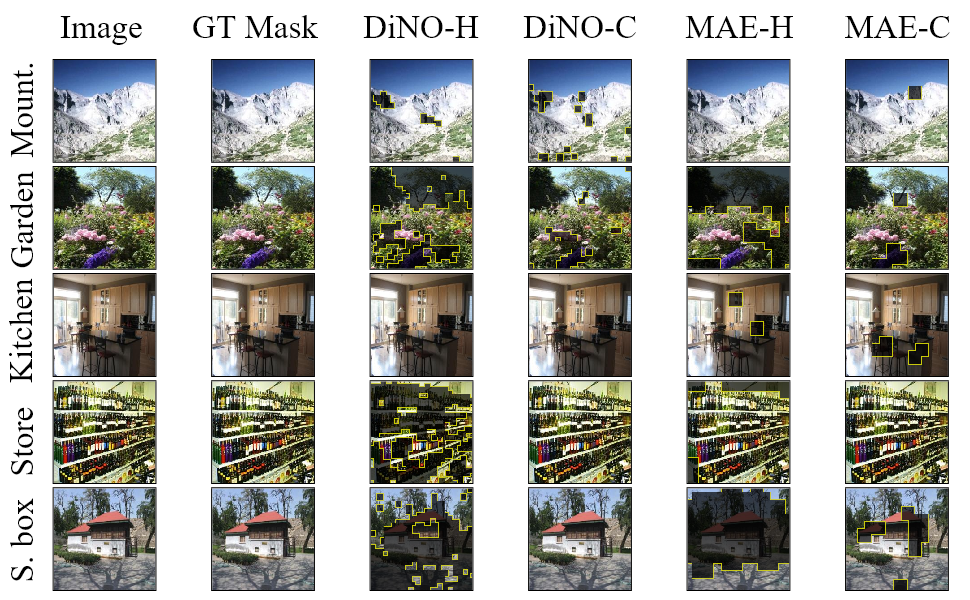}
        \caption{\scriptsize Scene: Snowy mountain, cottage garden, kitchen, liquor store, signal box.}
        \label{fig:sub4}
    \end{subfigure}
    
    \vspace{0.5cm} %
    
    \begin{subfigure}[t]{0.49\linewidth}
        \includegraphics[width=\linewidth]{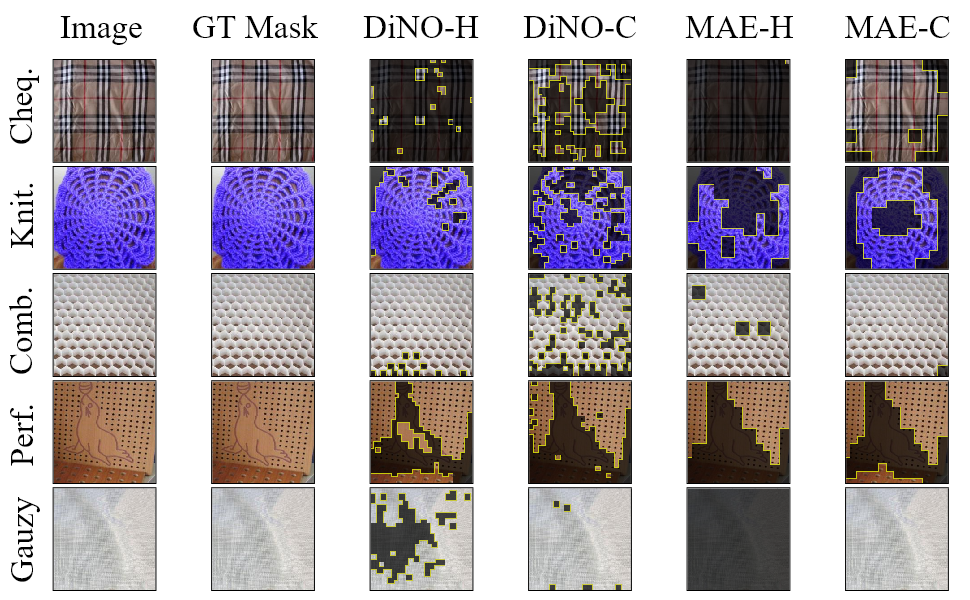}
        \caption{\scriptsize Texture: Chequered, knitted, honeycombed, perforated, gauzy.}
        \label{fig:sub5}
    \end{subfigure}
    \caption{Segmentation visualizations for DiNO and MAE templates using cosine (DiNO-C, MAE-C) and hyperplane (DiNO-H, MAE-H) decision rules.  Each figure showcases five unseen images from a specific concept category (material, object, part, scene, texture)}
    \label{fig:aggregated_results1}
\end{figure}

In Fig.~\ref{fig:aggregated_results1}, we present image samples organized by their primary category 
(material, object, part, scene, texture). Within each category, we select five representative concepts, 
and examine one image sample per concept. The selected concepts are chosen to ensure a balanced representation 
of 
DiNO hyperplane template performance, incorporating both the highest and lowest F1 scores. To improve 
visualization clarity, the representative image is selected from the test set based on the largest area 
coverage of the corresponding concept. Additionally, for each image sample, we provide its ground truth 
segmentation mask (GT Mask) alongside the predicted masks generated by hyperplane-based (DiNO-C, MAE-C) and 
cosine decision rule-based (DiNO-C, MAE-C) template models.
 
\begin{figure}[h!]
    \centering
    \begin{subfigure}[b]{0.49\linewidth}
        \includegraphics[width=\linewidth]{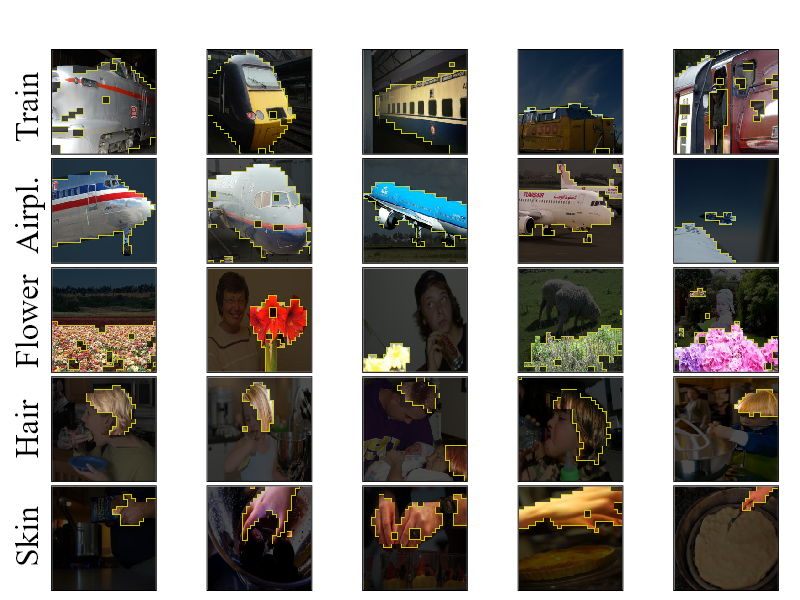}
        \caption{ \scriptsize DiNO hyperplane rule: Train, airplane, flower, hair, skin.}
        \label{fig:sub1}
    \end{subfigure}
    \begin{subfigure}[b]{0.49\linewidth}
        \includegraphics[width=\linewidth]{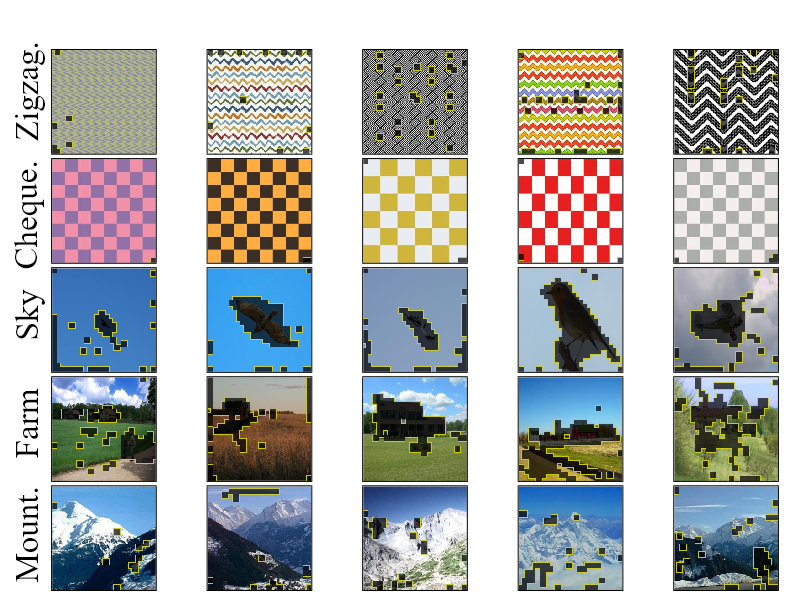}
        \caption{\scriptsize DiNO cosine rule: Zigzag, chequered, sky, farm, mountain}
        \label{fig:sub2}
    \end{subfigure}
     \begin{subfigure}[b]{0.49\linewidth}
        \includegraphics[width=\linewidth]{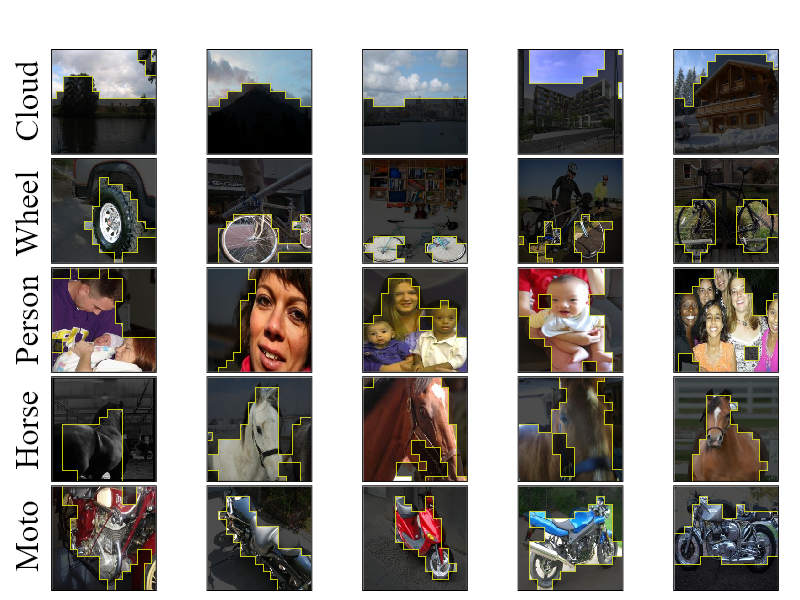}
        \caption{\scriptsize MAE hyperplane rule: Cloud, wheel, person, horse, motorcycle.}
        \label{fig:sub3}
    \end{subfigure}
     \begin{subfigure}[b]{0.49\linewidth}
        \includegraphics[width=\linewidth]{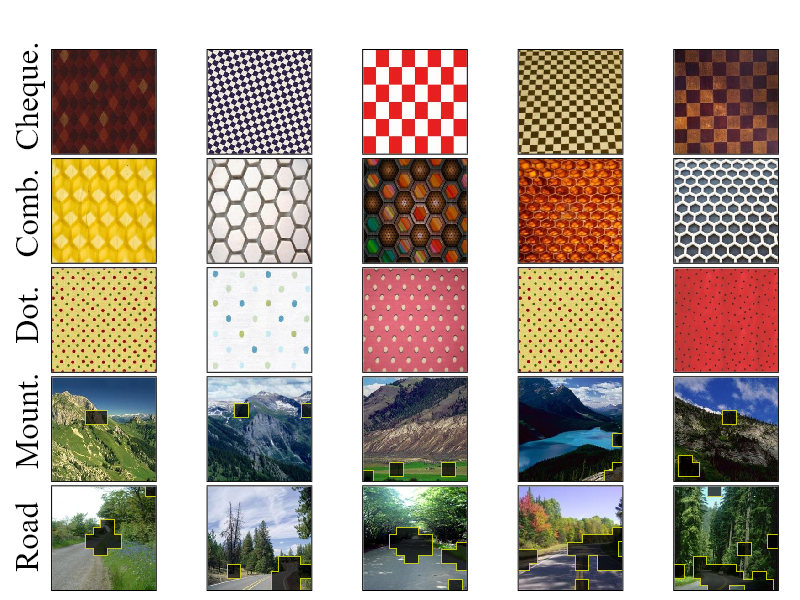}
        \caption{\scriptsize MAE cosine rule: Chequered, honeycombed, dotted, mountain, forest road}
        \label{fig:sub4}
    \end{subfigure}
    \caption{Segmentation visualizations for DiNO and MAE utilizing cosine and hyperplane decision rules. Each figure presents segmentation masks of unseen samples, produced by a particular model (DiNO, MAE) and decision rule (cosine, hyperplane). We showcase five samples per concept, highlighting the top five concepts with the highest F1 scores.}
    \label{fig:aggregated_results2}
\end{figure}

 In Fig.~\ref{fig:aggregated_results2}, we present segmentation visualizations for the concept labels with the highest F1 scores. Specifically, for each model (DiNO, MAE) and decision rule (hyperplane, cosine), we identify the top-five concept labels based on their F1 scores. For each selected concept, we showcase segmentation masks for five image samples where the template achieves the highest intersection over union (IoU).

\subsection{Summary}
Our post-hoc concept direction analysis provides insights into the 
representation power of pretrained DiNO and MAE models, offering guidelines for practical 
applications while raising questions for future work. A key observation is that the 
hyperplane classification rule consistently delivers better semantic separability than 
the cosine counterpart in both classification and segmentation downstream tasks. While 
MAE's [CLS] tokens seem to be an exception to this finding, we demonstrated that cosine 
distance between tokens is a suboptimal intra-class similarity metric.

Additionally, we showed that depending on the downstream task, context, and pretraining objective, different ViT tokens --some of which had not been extensively explored in the literature-- yield better semantic separability. This challenges current intuitions regarding the interpretation of query, key, and value tokens within transformer architectures and highlights the importance of understanding the role of each block within a transformer layer.

Furthermore, when  utilizing pretrained DiNO and MAE models in downstream tasks, the following observations should be mentioned: For image classification, DiNO's $\x2$ 
token combined with the hyperplane classification rule results in optimal 
classification results. Respectively, MAE's tokens should not be considered in 
this context as they produce random image classifiers. 
When labels are sparse and a few-shot context is required, 
DiNO's $\x1$ is better aligned with the cosine 
classification rule compared to other token types. 
We also observe that a support set size of $50$ samples 
represents the point at which performance gains begin to significantly diminish.
\par 
For semantic segmentation tasks, the models achieve their highest 
scores when leveraging their respective $\k$ tokens and the 
hyperplane decision rule. While DiNO outperforms MAE, the latter's 
strong performance in this context highlights that masked image 
modeling could serve as an important pretext (sub)task in the 
development of foundational vision transformers. Furthermore, 
DiNO's $\k$ tokens achieve the highest performance across all object 
categories, a trend that's not evident in MAE. Finally, in a few-shot context, both models' overall performance 
across all concept categories is inadequate. However, the $\q$ token 
for both DiNO and MAE provides excellent separability for textural 
concepts.

%% file: sections/conclusion.tex
\section{Limitations}
While our study provides a thorough analysis of self-supervised ViT properties across various pre-training 
objectives, token types, decision rules, downstream tasks, and contexts, it has certain limitations. A primary 
constraint was computational resources, which restricted our evaluation solely to ViT tokens extracted from the 
final transformer layer. Additionally, we treat image segmentation as a non-overlapping patch-level classification rather 
than pixel-level classification. Since ViT-based segmentation methods using frozen backbones \cite{stego} 
perform spatial interpolation of the feature maps to restore the spatial dimensionality of the input space prior 
to classification, our approach does not significantly deviate from this norm. Finally, regarding classification 
via the cosine decision rule, we did not account for feature-space centering prior to the computation of 
cosine-similarity between features. While it would be interesting to investigate its effects, we will consider it in 
future works. 

\section{Conclusion}
 
Our work conducted an in-depth post-hoc concept direction analysis to evaluate the representational power of pretrained DiNO and MAE token types in classification and segmentation downstream tasks. We examined their performance in both standard and few-shot learning contexts, utilizing hyperplane and cosine-similarity decision rules. Our findings show that the cosine decision rule --often used in unsupervised learning approaches-- consistently results in inferior semantic separability compared to its hyperplane counterpart. We also demonstrate that the optimal token type selection is highly dependent on these factors, while confirming that masked modeling effectively constructs competent backbones for image segmentation tasks.
 
Future research toward the development of foundational vision architectures should focus on deepening our understanding and interpretation of ViT tokens (arising from the unintuitive and possibly unexpected efficiency of key and query tokens, disproving the hypothesis that value tokens possess superiority), as well as assessing the efficacy of transformer layers, particularly under self-supervised pretraining objectives. Additionally, in unsupervised learning applications --where the cosine distance between ViT tokens is commonly used as an intra-class similarity metric-- exploring semantic proximity metrics beyond cosine similarity could enhance downstream task performance. Alternatively, a possible future research direction could be to work towards pre-training methods that will enforce interpretable concept alignment through the cosine rule, offering imminent enhancement of many existing unsupervised works that rely on a self-supervised backbone.

\subsubsection*{Acknowledgement}
This work has been supported by the EC funded Horizon Europe Framework Programme: CAVAA Grant Agreement 101071178.

%% file: main.bbl
\begin{thebibliography}{10}
\providecommand{\url}[1]{\texttt{#1}}
\providecommand{\urlprefix}{URL }
\providecommand{\doi}[1]{https://doi.org/#1}

\bibitem{Amir2021DeepVF}
Amir, S., Gandelsman, Y., Bagon, S., Dekel, T.: Deep vit features as dense visual descriptors. ArXiv  \textbf{abs/2112.05814} (2021)

\bibitem{beit}
Bao, H., Dong, L., Piao, S., Wei, F.: Beit: Bert pre-training of image transformers (2022), \url{https://arxiv.org/abs/2106.08254}

\bibitem{netdissect}
Bau, D., Zhou, B., Khosla, A., Oliva, A., Torralba, A.: Network dissection: Quantifying interpretability of deep visual representations. 2017 IEEE Conference on Computer Vision and Pattern Recognition (CVPR) pp. 3319--3327 (2017)

\bibitem{Bell2014IntrinsicII}
Bell, S., Bala, K., Snavely, N.: Intrinsic images in the wild. ACM Transactions on Graphics (TOG)  \textbf{33},  1 -- 12 (2014)

\bibitem{foundation_models}
Bommasani, R., Hudson, D.A., Adeli, E., Altman, R., Arora, S., Arx, S.v., Bernstein, M.S., Bohg, J., Bosselut, A., Brunskill, E., Brynjolfsson, E., Buch, S., Card, D., Castellon, R., Chatterji, N., Chen, A., Creel, K., Davis, J.Q., Demszky, D., Donahue, C., Doumbouya, M., Durmus, E., Ermon, S., Etchemendy, J., Ethayarajh, K., Fei-Fei, L., Finn, C., Gale, T., Gillespie, L., Goel, K., Goodman, N., Grossman, S., Guha, N., Hashimoto, T., Henderson, P., Hewitt, J., Ho, D.E., Hong, J., Hsu, K., Huang, J., Icard, T., Jain, S., Jurafsky, D., Kalluri, P., Karamcheti, S., Keeling, G., Khani, F., Khattab, O., Koh, P.W., Krass, M., Krishna, R., Kuditipudi, R., Kumar, A., Ladhak, F., Lee, M., Lee, T., Leskovec, J., Levent, I., Li, X.L., Li, X., Ma, T., Malik, A., Manning, C.D., Mirchandani, S., Mitchell, E., Munyikwa, Z., Nair, S., Narayan, A., Narayanan, D., Newman, B., Nie, A., Niebles, J.C., Nilforoshan, H., Nyarko, J., Ogut, G., Orr, L., Papadimitriou, I., Park, J.S., Piech, C., Portelance, E., Potts, C., Raghunathan,
  A., Reich, R., Ren, H., Rong, F., Roohani, Y., Ruiz, C., Ryan, J., Ré, C., Sadigh, D., Sagawa, S., Santhanam, K., Shih, A., Srinivasan, K., Tamkin, A., Taori, R., Thomas, A.W., Tramèr, F., Wang, R.E., Wang, W., Wu, B., Wu, J., Wu, Y., Xie, S.M., Yasunaga, M., You, J., Zaharia, M., Zhang, M., Zhang, T., Zhang, X., Zhang, Y., Zheng, L., Zhou, K., Liang, P.: On the opportunities and risks of foundation models (arXiv:2108.07258) (Jul 2022). \doi{10.48550/arXiv.2108.07258}, \url{http://arxiv.org/abs/2108.07258}, arXiv:2108.07258 [cs]

\bibitem{dino}
Caron, M., Touvron, H., Misra, I., J\'egou, H., Mairal, J., Bojanowski, P., Joulin, A.: Emerging properties in self-supervised vision transformers. In: Proceedings of the IEEE/CVF International Conference on Computer Vision (ICCV). pp. 9650--9660 (October 2021)

\bibitem{simclr}
Chen, T., Kornblith, S., Norouzi, M., Hinton, G.: A simple framework for contrastive learning of visual representations. In: International conference on machine learning. pp. 1597--1607. PMLR (2020)

\bibitem{pali}
Chen, X., Wang, X., Changpinyo, S., Piergiovanni, A.J., Padlewski, P., Salz, D.M., Goodman, S., Grycner, A., Mustafa, B., Beyer, L., Kolesnikov, A., Puigcerver, J., Ding, N., Rong, K., Akbari, H., Mishra, G., Xue, L., Thapliyal, A.V., Bradbury, J., Kuo, W., Seyedhosseini, M., Jia, C., Ayan, B.K., Riquelme, C., Steiner, A., Angelova, A., Zhai, X., Houlsby, N., Soricut, R.: Pali: A jointly-scaled multilingual language-image model. ArXiv  \textbf{abs/2209.06794} (2022)

\bibitem{Chen2014DetectWY}
Chen, X., Mottaghi, R., Liu, X., Fidler, S., Urtasun, R., Yuille, A.L.: Detect what you can: Detecting and representing objects using holistic models and body parts. 2014 IEEE Conference on Computer Vision and Pattern Recognition pp. 1979--1986 (2014)

\bibitem{moco_1}
Chen, X., Fan, H., Girshick, R., He, K.: Improved baselines with momentum contrastive learning (arXiv:2003.04297) (Mar 2020). \doi{10.48550/arXiv.2003.04297}, \url{http://arxiv.org/abs/2003.04297}, arXiv:2003.04297 [cs]

\bibitem{moco}
Chen, X., Xie, S., He, K.: An empirical study of training self-supervised vision transformers. 2021 IEEE/CVF International Conference on Computer Vision (ICCV) pp. 9620--9629 (2021)

\bibitem{Cimpoi2013DescribingTI}
Cimpoi, M., Maji, S., Kokkinos, I., Mohamed, S., Vedaldi, A.: Describing textures in the wild. 2014 IEEE Conference on Computer Vision and Pattern Recognition pp. 3606--3613 (2013)

\bibitem{cunningham2023sparse}
Cunningham, H., Ewart, A., Riggs, L., Huben, R., Sharkey, L.: Sparse autoencoders find highly interpretable features in language models. arXiv preprint arXiv:2309.08600  (2023)

\bibitem{vit}
Dosovitskiy, A., Beyer, L., Kolesnikov, A., Weissenborn, D., Zhai, X., Unterthiner, T., Dehghani, M., Minderer, M., Heigold, G., Gelly, S., Uszkoreit, J., Houlsby, N.: An image is worth 16x16 words: Transformers for image recognition at scale. ArXiv  \textbf{abs/2010.11929} (2020)

\bibitem{doumanoglou2023unsupervised}
Doumanoglou, A., Asteriadis, S., Zarpalas, D.: Unsupervised interpretable basis extraction for concept--based visual explanations. IEEE Transactions on Artificial Intelligence  (2023)

\bibitem{fel2023craft}
Fel, T., Picard, A., Bethune, L., Boissin, T., Vigouroux, D., Colin, J., Cad{\`e}ne, R., Serre, T.: Craft: Concept recursive activation factorization for explainability. In: Proceedings of the IEEE/CVF Conference on Computer Vision and Pattern Recognition. pp. 2711--2721 (2023)

\bibitem{stego}
Hamilton, M., Zhang, Z., Hariharan, B., Snavely, N., Freeman, W.T.: Unsupervised semantic segmentation by distilling feature correspondences. ArXiv  \textbf{abs/2203.08414} (2022)

\bibitem{vit-survey1}
Han, K., Wang, Y., Chen, H., Chen, X., Guo, J., Liu, Z., Tang, Y., Xiao, A., Xu, C., Xu, Y., et~al.: A survey on vision transformer. IEEE transactions on pattern analysis and machine intelligence  \textbf{45}(1),  87--110 (2022)

\bibitem{mae}
He, K., Chen, X., Xie, S., Li, Y., Doll\'ar, P., Girshick, R.: Masked autoencoders are scalable vision learners. In: Proceedings of the IEEE/CVF Conference on Computer Vision and Pattern Recognition (CVPR). pp. 16000--16009 (June 2022)

\bibitem{enchancer}
Kang, D., Koniusz, P., Cho, M., Murray, N.: Distilling self-supervised vision transformers for weakly-supervised few-shot classification \& segmentation. 2023 IEEE/CVF Conference on Computer Vision and Pattern Recognition (CVPR) pp. 19627--19638 (2023)

\bibitem{vit-survey3}
Khan, S., Naseer, M., Hayat, M., Zamir, S.W., Khan, F.S., Shah, M.: Transformers in vision: A survey. ACM computing surveys (CSUR)  \textbf{54}(10s),  1--41 (2022)

\bibitem{imagenettiny}
Le, Y., Yang, X.S.: Tiny imagenet visual recognition challenge (2015)

\bibitem{vit-survey2}
Liu, Y., Zhang, Y., Wang, Y., Hou, F., Yuan, J., Tian, J., Zhang, Y., Shi, Z., Fan, J., He, Z.: A survey of visual transformers. IEEE Transactions on Neural Networks and Learning Systems  (2023)

\bibitem{Mishra2022ASE}
Mishra, S.K., Robinson, J., Chang, H., Jacobs, D., Sarna, A., Maschinot, A., Krishnan, D.: A simple, efficient and scalable contrastive masked autoencoder for learning visual representations. ArXiv  \textbf{abs/2210.16870} (2022)

\bibitem{Mottaghi2014TheRO}
Mottaghi, R., Chen, X., Liu, X., Cho, N.G., Lee, S.W., Fidler, S., Urtasun, R., Yuille, A.L.: The role of context for object detection and semantic segmentation in the wild. 2014 IEEE Conference on Computer Vision and Pattern Recognition pp. 891--898 (2014)

\bibitem{vit-intriguing}
Naseer, M.M., Ranasinghe, K., Khan, S.H., Hayat, M., Shahbaz~Khan, F., Yang, M.H.: Intriguing properties of vision transformers. Advances in Neural Information Processing Systems  \textbf{34},  23296--23308 (2021)

\bibitem{Nguyen2024ExploringSV}
Nguyen, H.H., Yamagishi, J., Echizen, I.: Exploring self-supervised vision transformers for deepfake detection: A comparative analysis. ArXiv  \textbf{abs/2405.00355} (2024)

\bibitem{dinov2}
Oquab, M., Darcet, T., Moutakanni, T., Vo, H.Q., Szafraniec, M., Khalidov, V., Fernandez, P., Haziza, D., Massa, F., El-Nouby, A., Assran, M., Ballas, N., Galuba, W., Howes, R., Huang, P.Y.B., Li, S.W., Misra, I., Rabbat, M.G., Sharma, V., Synnaeve, G., Xu, H., J{\'e}gou, H., Mairal, J., Labatut, P., Joulin, A., Bojanowski, P.: Dinov2: Learning robust visual features without supervision. ArXiv  \textbf{abs/2304.07193} (2023)

\bibitem{Park2023WhatDS}
Park, N., Kim, W., Heo, B., Kim, T., Yun, S.: What do self-supervised vision transformers learn? ArXiv  \textbf{abs/2305.00729} (2023)

\bibitem{beit2}
Peng, Z., Dong, L., Bao, H., Ye, Q., Wei, F.: Beit v2: Masked image modeling with vector-quantized visual tokenizers (arXiv:2208.06366) (Oct 2022). \doi{10.48550/arXiv.2208.06366}, \url{http://arxiv.org/abs/2208.06366}, arXiv:2208.06366 [cs]

\bibitem{mae-probe}
Qian, Y., Wang, Y., Lin, J.: Enhancing the linear probing performance of masked auto-encoders. In: International Conference on Pattern Recognition. pp. 289--301. Springer (2022)

\bibitem{Radford2021LearningTV}
Radford, A., Kim, J.W., Hallacy, C., Ramesh, A., Goh, G., Agarwal, S., Sastry, G., Askell, A., Mishkin, P., Clark, J., Krueger, G., Sutskever, I.: Learning transferable visual models from natural language supervision. In: International Conference on Machine Learning (2021), \url{https://api.semanticscholar.org/CorpusID:231591445}

\bibitem{vit-see}
Raghu, M., Unterthiner, T., Kornblith, S., Zhang, C., Dosovitskiy, A.: Do vision transformers see like convolutional neural networks? Advances in neural information processing systems  \textbf{34},  12116--12128 (2021)

\bibitem{rao2024discover}
Rao, S., Mahajan, S., B{\"o}hle, M., Schiele, B.: Discover-then-name: Task-agnostic concept bottlenecks via automated concept discovery. In: European Conference on Computer Vision. pp. 444--461. Springer (2024)

\bibitem{imagenet}
Russakovsky, O., Deng, J., Su, H., Krause, J., Satheesh, S., Ma, S., Huang, Z., Karpathy, A., Khosla, A., Bernstein, M.S., Berg, A.C., Fei-Fei, L.: Imagenet large scale visual recognition challenge. International Journal of Computer Vision  \textbf{115},  211 -- 252 (2014)

\bibitem{Lost}
Sim{\'e}oni, O., Puy, G., Vo, H.V., Roburin, S., Gidaris, S., Bursuc, A., P'erez, P., Marlet, R., Ponce, J.: Localizing objects with self-supervised transformers and no labels. ArXiv  \textbf{abs/2109.14279} (2021)

\bibitem{deit}
Touvron, H., Cord, M., Douze, M., Massa, F., Sablayrolles, A., J'egou, H.: Training data-efficient image transformers \& distillation through attention. In: International Conference on Machine Learning (2020)

\bibitem{Vanyan2023AnalyzingLR}
Vanyan, A., Barseghyan, A., Tamazyan, H., Huroyan, V., Khachatrian, H., Danelljan, M.: Analyzing local representations of self-supervised vision transformers. ArXiv  \textbf{abs/2401.00463} (2023)

\bibitem{one_peace}
Wang, P., Wang, S., Lin, J., Bai, S., Zhou, X., Zhou, J., Wang, X., Zhou, C.: One-peace: Exploring one general representation model toward unlimited modalities. ArXiv  \textbf{abs/2305.11172} (2023)

\bibitem{beit3}
Wang, W., Bao, H., Dong, L., Bjorck, J., Peng, Z., Liu, Q., Aggarwal, K., Mohammed, O.K., Singhal, S., Som, S., Wei, F.: Image as a foreign language: Beit pretraining for all vision and vision-language tasks. ArXiv  \textbf{abs/2208.10442} (2022)

\bibitem{dense_clr}
Wang, X., Zhang, R., Shen, C., Kong, T., Li, L.: Dense contrastive learning for self-supervised visual pre-training. In: 2021 IEEE/CVF Conference on Computer Vision and Pattern Recognition (CVPR). p. 3023–3032. IEEE, Nashville, TN, USA (Jun 2021). \doi{10.1109/CVPR46437.2021.00304}, \url{https://ieeexplore.ieee.org/document/9578497/}

\bibitem{cutler}
Wang, X., Girdhar, R., Yu, S.X., Misra, I.: Cut and learn for unsupervised object detection and instance segmentation. 2023 IEEE/CVF Conference on Computer Vision and Pattern Recognition (CVPR) pp. 3124--3134 (2023)

\bibitem{tokencut}
Wang, Y., Shen, X., Hu, S.X., Yuan, Y., Crowley, J.L., Vaufreydaz, D.: Self-supervised transformers for unsupervised object discovery using normalized cut. 2022 IEEE/CVF Conference on Computer Vision and Pattern Recognition (CVPR) pp. 14523--14533 (2022)

\bibitem{self_supervised}
Wu, H., Gao, Y., Zhang, Y., Lin, S., Xie, Y., Sun, X., Li, K.: Self-supervised models are good teaching assistants for vision transformers. In: Proceedings of the 39th International Conference on Machine Learning. p. 24031–24042. PMLR (Jun 2022), \url{https://proceedings.mlr.press/v162/wu22c.html}

\bibitem{simim}
Xie, Z., Zhang, Z., Cao, Y., Lin, Y., Bao, J., Yao, Z., Dai, Q., Hu, H.: Simmim: a simple framework for masked image modeling. 2022 IEEE/CVF Conference on Computer Vision and Pattern Recognition (CVPR) pp. 9643--9653 (2021)

\bibitem{efficientsam}
Xiong, Y., Varadarajan, B., Wu, L., Xiang, X., Xiao, F., Zhu, C., Dai, X., Wang, D., Sun, F., Iandola, F., et~al.: Efficientsam: Leveraged masked image pretraining for efficient segment anything. In: Proceedings of the IEEE/CVF Conference on Computer Vision and Pattern Recognition. pp. 16111--16121 (2024)

\bibitem{Yang2024DepthAU}
Yang, L., Kang, B., Huang, Z., Xu, X., Feng, J., Zhao, H.: Depth anything: Unleashing the power of large-scale unlabeled data. 2024 IEEE/CVF Conference on Computer Vision and Pattern Recognition (CVPR) pp. 10371--10381 (2024)

\bibitem{coca}
Yu, J., Wang, Z., Vasudevan, V., Yeung, L., Seyedhosseini, M., Wu, Y.: Coca: Contrastive captioners are image-text foundation models. Trans. Mach. Learn. Res.  \textbf{2022} (2022)

\bibitem{Yue2023UnderstandingMA}
Yue, X., Bai, L., Wei, M., Pang, J., Liu, X., Zhou, L., Ouyang, W.: Understanding masked autoencoders from a local contrastive perspective. ArXiv  \textbf{abs/2310.01994} (2023)

\bibitem{Zhang2023MultimodalFA}
Zhang, W., Ma, B., Qiu, F., qiong Ding, Y.: Multi-modal facial affective analysis based on masked autoencoder. 2023 IEEE/CVF Conference on Computer Vision and Pattern Recognition Workshops (CVPRW) pp. 5793--5802 (2023)

\bibitem{Zhou2018InterpretableBD}
Zhou, B., Sun, Y., Bau, D., Torralba, A.: Interpretable basis decomposition for visual explanation. In: European Conference on Computer Vision (2018)

\bibitem{Zhou2017ScenePT}
Zhou, B., Zhao, H., Puig, X., Fidler, S., Barriuso, A., Torralba, A.: Scene parsing through ade20k dataset. 2017 IEEE Conference on Computer Vision and Pattern Recognition (CVPR) pp. 5122--5130 (2017)

\bibitem{mae-medical}
Zhou, L., Liu, H., Bae, J., He, J., Samaras, D., Prasanna, P.: Self pre-training with masked autoencoders for medical image classification and segmentation (arXiv:2203.05573) (Apr 2023). \doi{10.48550/arXiv.2203.05573}, \url{http://arxiv.org/abs/2203.05573}, arXiv:2203.05573 [eess]

\end{thebibliography}
